\documentclass{article}

\usepackage{booktabs}
\usepackage{PRIMEarxiv}
\usepackage{tcolorbox}
\tcbuselibrary{breakable}
\usepackage[utf8]{inputenc} 
\usepackage[T1]{fontenc}    
\usepackage{hyperref}       
\usepackage{url}            
\usepackage{booktabs}       
\usepackage{amsfonts}       
\usepackage{nicefrac}       
\usepackage{microtype}      
\usepackage{lipsum}
\usepackage{fancyhdr}       
\usepackage{graphicx}       
\usepackage{xcolor}         
\usepackage{array}          
\usepackage{amsmath}        
\usepackage{tikz}
\usetikzlibrary{positioning, shapes, arrows, fit, backgrounds, calc}
\usepackage{xcolor}
\usepackage{float}
\usepackage[minnames=6,maxnames=6,sorting=none,backend=bibtex]{biblatex}
\graphicspath{{media/}}     

\title{Deep Research Bench:\\Evaluating AI Web Research Agents
\thanks{
  Authors are listed in alphabetical order.}
}

\author{
  \And
  {\hspace{17em}}
  FutureSearch* \\
  \And
  {\hspace{17em}}
  \And
  {\hspace{6em}}
  \And
  Nikos I. Bosse\\
  \And
  Jon Evans\\
  \And
  Robert G. Gambee\\
  \And
  Daniel Hnyk\\
  \And
  {\hspace{6em}}
  \And
  {\hspace{5em}}
  \And
  Peter M\"{u}hlbacher\\
  \And
  Lawrence Phillips\\
  \And
  Dan Schwarz\\
  \And
  Jack Wildman \\
  \And
  {\hspace{5em}}
}

\newcommand{\exampletaskbox}[2]{%
\vspace{1em}
\begin{tcolorbox}[
    breakable,
    width=\textwidth,
    left=10pt,
    right=10pt,
    top=6pt,
    bottom=6pt,
    boxrule=0.5pt
]
\textbf{Example instance: #1}\vspace{0.5em}\\
#2
\end{tcolorbox}
}

\begin{document}
\maketitle

\begin{abstract}
  Amongst the most common use cases of modern AI is LLM chat with web search enabled. However, no direct evaluations of the quality of web research agents exist that control for the continually-changing web. We introduce Deep Research Bench, consisting of 89 multi-step web research task instances of varying difficulty across 8 diverse task categories, with the answers carefully worked out by skilled humans. We provide a "RetroSearch" environment with a large frozen set of scraped web pages, and demonstrate that offline "RetroSearch" agents perform comparably to "live web" agents, enabling reliable evaluations of models over time. We provide robust agent tooling and scaffolding to benchmark major LLMs as they are released, including "thinking" models like o3 and Gemini 2.5 Pro. We include automated evaluations of the lengthy agent traces to report progress over time in hallucinations, tool use, and forgetting. Finally, we evaluate the major web research products branded as "Deep Research", "Deep Search", "Search", or "Research." Results are available on a public leaderboard at \url{https://drb.futuresearch.ai/}.
\end{abstract}


\section{Introduction}
Large Language Models (LLMs) are presently used to power a wide range of applications, from chatbots to search engines to code editors. Agents based on LLMs can solve complex problems and are becoming increasingly useful for day-to-day tasks. As technologies continue to advance, there is increased demand to understand their capabilities, limitation, and change over time.

Benchmarks provide quantitative metrics that are easy to compare, replicable, and ideally objective. Benchmarks that reflect performance on real-world tasks can provide useful insights to researchers, industry practitioners, and policymakers to navigate inherent uncertainties, as well as to design policies and regulatory frameworks.

We present Deep Research Bench, a benchmark designed to evaluate LLM agents on complex, messy real-world web research tasks. These tasks reflect the kinds of research problems that analysts in a variety of industries encounter and for which they use AI web research agents. We present the results from running eleven commercial web research products (such as OpenAI Deep Research), as well as ReAct agents powered by eleven models (such as o3, Claude Sonnet 3.7).

Many benchmarks evaluate LLMs. Fewer evaluate LLM agents. Earlier testing frameworks such as WebShop \cite{yaoWebShopScalableRealWorld2023}, WebArena \cite{zhouWebArenaRealisticWeb2024}, Jericho \cite{hausknechtInteractiveFictionGames2020}, AgentSims \cite{linAgentSimsOpenSourceSandbox2023} or ALFWorld \cite{shridharALFWorldAligningText2021} (a set of virtual household tasks) focused on testing the capabilities of LLMs in a controlled environment. While testing a wide range of capabilities, the problems agents have to solve are often somewhat detached from the kinds of messy and complicated real-world tasks humans have to solve in their day-to-day lives. And those that do, such as AgentBoard, do not generally involve the web.

Two benchmarks that do involve the open web are AgentBench and GAIA. AgentBench \cite{liuAgentBenchEvaluatingLLMs2023} is one of the earliest benchmarks, and handles three main domains: web, code, and games (partly making use of previously existing testing frameworks such as Web Shop \cite{yaoWebShopScalableRealWorld2023} or ALFWord \cite{shridharALFWorldAligningText2021} and therefore suffering from similar limitations). AgentBench has not been updated to show performance of models published after 2023.
GAIA \cite{mialonGAIABenchmarkGeneral2023} goes beyond AgentBench by featuring tasks that require browsing the open web, and testing the ability of agents to handle multi-modal inputs. It evaluates a wide range of capabilities such as reasoning, coding, web browsing, and handling tools. Tasks are separated into three different difficulty levels, but are somewhat limited in the sense that even on the highest difficulty level, humans are able to solve around 90\% of the tasks in about 20 minutes. The tasks' answers are such that crowdworkers can derive them with confidence in their objective correctness (this is how they're constructed), and are designed to be stable over time (although their stability is not guaranteed). Due to these constraints, the tasks are often somewhat contrived, failing to translate directly into a measure of economically valuable labour. Furthermore, tasks suffer from time decay as information may appear or disappear from the web.

Deep Research Bench extends our previous work \cite{muhlbacherRealisticLongTermBenchmark2024} and is a benchmark designed to remove task degradation and random variation of the web, while still evaluating LLM agents on complex, messy real-world web research tasks. Deep Research Bench uses RetroSearch, an environment for querying a large database of websites scraped from the open web. This avoids issues when the ground truth of tasks change: when new information appears, or when information becomes more salient, e.g. when someone publishes a new analysis summarizing pre-existing information.

Furthermore, a large part of the tasks are informed by FutureSearch's work as a startup providing AI-centred research and intelligence to clients, rather than relying on artificial scenarios. The tasks that make up Deep Research Bench are also used internally by FutureSearch to evaluate and iterate on our own research tools.

We commit to continuously updating Deep Research Bench and the public leaderboard to include new tasks, as well as running new models on Deep Research Bench as they are released. Over time, we expect the portion of instances that came from our own, for-profit research to grow further. Evaluations are available at \url{drb.futuresearch.ai/}. The Appendix provides a sample of eight task instances, yet we do not publish the full set of instances to avoid data contamination.

In Section \ref{sec:methods}, we provide details on the basic setup of the benchmark and the evaluations. In Section \ref{sec:results}, we present a quantitative and qualitative analysis of the performance of LLM agents and the commercial web research products on Deep Research Bench. In Section \ref{sec:discussion}, we discuss our findings and potential future work.

\section{Methods}
\label{sec:methods}

In this Section, we describe the tasks and instances in detail (\ref{sec:task-instances}), provide a subjective assessment of difficulty and techniques demanded by tasks and instances(\ref{sec:subjective-assessment}), explain our shallow-elicitation approach to prompting each stage of the agent loop (\ref{sec:prompting}), give details on the agents, architectures, and LLMs (\ref{sec:agents-tools-llms}) as well as the commercial web research products we evaluated (\ref{sec:methods-web-research-products}), provide information on how we ran the evaluations (\ref{sec:running-the-evaluation}), present RetroSearch, our custom API that serves agents a frozen version of the web (\ref{sec:retrosearch}), and provide details on how we evaluate performance (\ref{sec:methods-evaluating-performance}).

\subsection{Benchmark Tasks}
\label{sec:task-instances}

Deep Research Bench consists of eight tasks, comprising a total set of 89 instances, representing atomized versions of problems researchers or analysts encounter when researching a topic on the web. Table \ref{tab:task-type-instances} shows an overview.

In an iterative process, we use the research we do for clients to generate new tasks and instances. This ensures both that the tasks are realistic and valuable, and that the answers are correct. Those tasks, in turn, are used to evaluate and improve our tools and workflows. As of the publication of this paper, $\sim$40\% of instances have been generated from client work. We expect this to be the main source of new tasks instances.

Another $\sim$25\% of instances have been generated by making use of the in-depth research already extant on the web, and blocking that web source from the agents. Examples include data on AI models from Epoch AI \cite{EpochAI}, and in-depth analyses from Deloitte. Agents attempting to reproduce these conclusions are an approximate simulation for the work done at these organizations.

The remaining $\sim$35\% of the instances were based neither on our research nor on that of others; we invented questions \textit{ex nihilo} that we thought were interesting, appropriately challenging, and --- perhaps with one Pokémon-related exception --- representative of real-world research.

Deep Research Bench is designed to have the following properties:

\begin{itemize}
  \item \textbf{Construct Validity:} Tasks arise from messy problems in real-world professions
  \item \textbf{Consequential:} Web agents that can perform these tasks well would be economically valuable
  \item \textbf{Continuous Scale of Performance:} The benchmark can identify quality differences between agents of similar web research capabilities
\end{itemize}

The tasks and their instances are summarized in Table \ref{tab:task-type-instances}.

\begin{table}[h]
  \centering
  \renewcommand{\arraystretch}{1.2}
  \begin{tabular}{p{2.7cm}>{\raggedright\arraybackslash}p{8.3cm}cp{2cm}}
    \toprule
    \textbf{Task Type} & \textbf{Description and Example} & \textbf{\# Inst.} & \textbf{Offline \newline Web Pages} \\
    \midrule
    Find Number & Find a reliable, known number on the internet. \vspace{0.1em} \newline \textit{The total number of FDA Class II Product Recalls of medical devices.} & 18 & 15,284 \newline (5k--39k) \\
    \addlinespace
    Find Dataset & Find links to datasets relevant to a given query. \vspace{0.1em} \newline \textit{How many IPOs with an offer price of at least \$5.00 were there in each year between 1980 and 2024?} & 12 & 20,627 \newline (10k--37k) \\
    \addlinespace
    Find Original \newline Source & Find the original source of a given claim. \vspace{0.1em} \newline \textit{From <LINK>, more than 8 out of 1000 users clicked on a phishing link monthly in 2024, up 190\% vs 2023.} & 11 & 38,518 \newline (15k--63k) \\
    \addlinespace
    Validate Claim & Estimate the probability that a given claim on the internet is true. \vspace{0.1em} \newline \textit{The median energy usage of ChatGPT queries is at least 10× greater than Google searches.} & 12 & 9,993 \newline (4k--22k) \\
    \addlinespace
    Derive Number & Derive a number not known on the internet, but derivable from known information. \vspace{0.1em} \newline \textit{How many IM and GM account closures did chess.com report for 2024?} & 10 & 17,042 \newline (6k--27k) \\
    \addlinespace
    Gather Evidence & Identify key pieces of evidence relevant to a given query. \vspace{0.1em} \newline \textit{What is the difficulty of the problems in the FrontierMath benchmark?} & 9 & 110,946 \newline (44k--189k) \\
    \addlinespace
    Populate Reference Class & Compile a list of instances that fit a given description. \vspace{0.1em} \newline \textit{List functioning satellites impacted by accidental high-speed collisions in space.} & 10 & 15,940 \newline (6k--35k) \\
    \addlinespace
    Compile Dataset & Compile a dataset based on a description of the desired data and required columns. \vspace{0.1em} \newline \textit{Software developer jobs in the US from 2019-2023 with columns: year, number, source, url, percent change.} & 7 & 15,233 \newline (5k--25k) \\
    \bottomrule
  \end{tabular}
  \vspace{0.5cm}
  \caption{Overview of task types. Offline Web Pages shows the median (min--max) number of web pages scraped and stored for agents to search and read during each task.}
  \label{tab:task-type-instances}
\end{table}

\subsection{Task and Instance Difficulty}
\label{sec:subjective-assessment}

In the following, we give a subjective assessment of tasks and the challenges humans face in solving them, finding that good solutions from agents broadly look like good solutions from humans. They usually involve these steps and challenges:

\begin{itemize}
  \item \textbf{Strategy formulation:} Good judgment on how the task might be solved, e.g. where to look for good information
  \item \textbf{Searching:} Formulating good search queries
  \item \textbf{Page selection:} Identifying the most promising and credible pages to read
  \item \textbf{Page reading:} Identifying useful information from pages
  \item \textbf{Strategy updating:} Given the information found, deciding what would next be most helpful
  \item \textbf{Reconciliation:} Synthesizing information, often contradictory, into the best overall view
  \item \textbf{Stopping:} Deciding whether to continue researching or stopping to give a final answer
\end{itemize}

We have developed a taxonomy of research and reasoning failures across these steps, available in Section \ref{sec:qualitative-assessment}. Here, we go into more detail on what challenges arise in each task category.

\subsubsection{Find Number}
Given a query for a single numeric value, an agent must find a reliable source for that number and return the number, the source, and the excerpt from the source which establishes that value.

A primary challenge in this task is that the agent encounters many misleading numbers --- sometimes straightforwardly wrong, non-credible numbers, but often credible-looking numbers that have a slightly different definition or are out of date.

This task evaluates an agent's ability to weight the credibility of disparate sources and, for some instances, its willingness to admit that the number in question cannot be attributed to any reliable source and concede failure.

\subsubsection{Find Dataset}
This task tests an agent's competence at formulating and iterating on targeted Google searches. For most instances, naive "keywordization" of the instance's query fails to find a page leading to one of the required datasets; careful choice of key terms, and in some cases site filters, is needed. Relatedly, performance here is (lightly) loaded on the ability to formulate and follow systematic research plans; it's sometimes helpful to do background reading on a field to learn about terms of art and/or respected data providers and analysts before searching directly for data.

Another capability under test here is judgment as to what constitutes a relevant result. For most instances, superficially relevant data is available and is easier to find than the directly relevant data our scoring requires, and agents must resist the temptation to terminate upon finding lower-quality results.

Finally, the task requires reasoning over a fairly long history, or competently using a scratchpad. Agents following a proper strategy will conduct a thorough search regardless of what they find early on (we can't know if we won't find something better until we look hard), and if high-quality data is found early, it must be recalled at the end of the session, avoiding distraction from lower-quality results found later on.

\subsubsection{Find Original Source}
Given a purported claim or fact, and its source, the agent is tasked with establishing the original source of that claim, often (when the claim is not spurious) a scientific paper or official dataset. It must return both the URL of the original source and the excerpt containing the original fact or claim.

This evaluates an agent's ability to traverse online information and find the root source of a cited claim. It not only requires investigating claim's history by citation or by time, but the ability to judge whether a source is genuinely original or itself just an interim reference to a further parent source.

\subsubsection{Validate Claim}
This task tests an agent's ability to not only collect relevant information, but also judge how well that evidence supports or refutes the claim. Is the evidence from a reliable source? When multiple sources agree, is that because they independently arrived at the same answer, or because they are echoing one another? When a source reports a number, what is the associated confidence interval, even if not explicitly stated? Additionally, the agent must exercise skepticism by applying its own priors. And at the end of its deliberation, the agent must distill all this qualitative and quantitative information into a numerical probability.

On top of this, claims require some amount of interpretation. For example, the claim "The median electrical energy usage of ChatGPT queries is at least 10 times greater than the median energy usage of Google searches" is relatively well-specified but still has room for interpretation. What types of energy consumption should be included? Do API requests count, or only those made via a web interface? What geographic regions and timeframes should be considered? To achieve optimal performance, an agent must recognize its assumptions and factor its uncertainties into its final answer.

\subsubsection{Derive Number}
This task tests an agent's ability to calculate or derive a number whose value can only be determined by intelligently combining values from multiple different sources. As such, performance here is to often downstream of Find Number's ability to correctly obtain such values.

At least as important, however, is an agent's ability to construct both a formula for the solution and a plan to acquire the terms for that formula; and, often crucially, its ability to refine and pivot its plan in case the initial one runs into difficulties, whether epistemic (e.g. inability to establish certain kinds of information) or practical (e.g. plans which require gathering and managing excessive numbers of data points.)

Derive Number is exponentially more difficult than Find Number, in that Derive almost always implies multiple Finds, so its base failure rate is Find Number's failure rate raised to the power of the number of terms to be found. As such the key to Derive Number success is often the ability to reduce this exponential decay, and minimize the number of required terms, with intelligent strategy.

\subsubsection{Gather Evidence}
The abilities under test here are broadly similar to those assessed by Find Dataset, but the instances are more challenging across all dimensions. Good performance requires formulating workable systematic research plans on the fly, and executing them competently. An instance might require the agent to enumerate the major lines of business of a company, and for each line, conduct a search for recent performance, when nowhere in the task prompt does it mention that the agent must do this.

The challenge of distinguishing between key evidence and related-but-not-informative information is also heightened here; in many instances there is abundant information about the overall topic readily available, but few key, directly relevant nuggets of evidence to find.

Another requirement for good performance, not present in Find Dataset, is the ability to recognize when and where additional context is needed in order for a given piece of evidence to be interpreted correctly.

\subsubsection{Populate Reference Class}
This task tests an agent's ability to identify a set of items that fit a given description. This requires coming up with an appropriate search strategy to identify potential candidates, as well as a way to verify that candidates are indeed valid members of the reference class and meet the inclusion criteria.

Sometimes, it is easiest to start with an exhaustive list of items and check the inclusion criteria one by one (e.g. to find a list of US Government Departments that have experienced cybersecurity breaches). Other times, it may be better to come up with a more targeted approach (e.g. to find a list of satellites that have been impacted by an accidental collision in space).

One fundamental challenge knowing when the task is done. For example, when compiling a list of incidents where pathogens have leaked from a lab, it is not clear what the total number is. Also, when forced to only return a certain number of items, agents must exercise judgement about which items are most relevant to the task.

Verifying whether or not a candidate item actually meets the inclusion criteria can be particularly challenging. Agents must not be misled by unreliable or outdated information.

Instances of this task can take humans hours to days to complete, and they often leave some uncertainty about whether they have overlooked some results. An agent that is able to score highly on this tasks should be considered a very capable research assistant that is able to perform complex research tasks in highly ambiguous environments.

\subsubsection{Compile Dataset}

This task tests an agent's ability to systematically collect and compile specific information from various different sources. The easier instances are akin to a collection of Find Number or Derive Number tasks.

More challenging instances involve compiling the dataset row by row, which is similar to the Populate Reference Class task. Others require starting from existing data, which depends on capability in Find Dataset plus the ability to combine information from existing datasets by filtering and manipulating the data appropriately.

This task also tests the ability to come up with a research plan and revise it as needed. For the most challenging task instances, the agent also has to exercise judgment about what data is relevant to the query and what is not. For example, when asking the agent to compile a dataset of the API costs for current top LLM models, it is not a priori clear what models to include. This poses similar challenges as the Populate Reference Class task, in addition to finding and combining the data.

\subsection{Prompting}\label{sec:prompting}
Many of the tasks comprising Deep Research Bench admit low-effort solutions that technically satisfy the task requirements, but are of much lower quality than the optimal answer.

For tasks like this, rather than explicitly cajoling models to give an optimal solution and providing details on what such a solution would look like, we prompt agents simply, generally in a few sentences, instructing them to solve the task and nothing more. This more accurately reflects the prompts that agents are likely to receive when asked to carry out research tasks in the real world, behind chat UIs.
It also allows us to investigate models' innate tendencies towards optimization or satisficement --- a question that has implications for AI safety.

This approach has drawbacks, however: it increases the degree of subjectivity inherent in our evaluation functions, and it leaves open the possibility that performance would be significantly better under different prompting.

In light of the latter point, the findings in this paper are to be interpreted as relevant only to low-elicitation prompting; we caveat claims with "under low elicitation" where appropriate. We will explore the high-elicitation case in future work. See examples of several-sentence elicitation prompts in the Appendix.

\subsection{Agents, Tools, and LLMs}
\label{sec:agents-tools-llms}

\subsubsection{Selection of LLMs and Products}

We aimed to evaluate a range of different models and providers, both open and closed, old models and new, and ensuring to include all the state-of-the-art models and products as of May 2025. These are presented in table \ref{tab:models-and-tools}.

\begin{table}[h]
  \renewcommand{\arraystretch}{1.2}
  \small
  \centering
  \begin{tabular}{llll}
    \toprule
    \textbf{Name} & \textbf{Spec} & \textbf{Provider} & \textbf{Weights}\\
    \midrule
    \multicolumn{4}{c}{\textbf{Thinking LLMs}} \\
    Claude Sonnet 3.7 (Thinking) & claude-3-7-sonnet-20250219 & Anthropic API + AWS Bedrock & Closed \\
    DeepSeek-R1 & deepseek-r1 & Openrouter & Open  \\
    Gemini 2.5 Pro & gemini-2.5-pro-preview-03-25 & Vertex AI & Closed \\
    Gemini 2.5 Flash (Thinking) & gemini-2.5-flash-preview-04-17 & Vertex AI & Closed \\
    o3 & o3-2025-04-16 & OpenAI API & Closed \\
    \midrule
    \multicolumn{4}{c}{\textbf{Non-thinking LLMs}} \\
    Claude Sonnet 3.7 & claude-3-7-sonnet-20250219 & Anthropic API + AWS Bedrock & Closed \\
    Gemini 2.5 Flash & gemini-2.5-flash-preview-04-17 & Vertex AI & Closed \\
    Gemma 3 & gemma-3-27b-it & Self-deployed on Vertex AI & Open \\
    GPT 4 & gpt-4-0613 & OpenAI API + Azure OpenAI & Closed \\
    GPT 4.1 & gpt-4.1-2025-04-14 & OpenAI API + Azure OpenAI & Closed \\
    GPT 4 Turbo & gpt-4-turbo-2024-04-09 & OpenAI API + Azure OpenAI & Closed \\
    Mistral Small & mistral-small-2503 & Mistral API (via Vertex) & Open \\
    \midrule
    \multicolumn{4}{c}{\textbf{Web Research Products}} \\
    OpenAI Deep Research & & OpenAI & \\
    ChatGPT o3 + Web Research & & OpenAI & \\
    ChatGPT 4.5 + Web Research & & OpenAI & \\
    Gemini Deep Research & & Google & \\
    Gemini 2.5 Pro + Search & & Google & \\
    Grok DeepSearch & & xAI & \\
    Perplexity Pro (best LLM) & & Perplexity & \\
    Perplexity Deep Research & & Perplexity & \\
    Claude Research + Extended Thinking & & Anthropic & \\
    Claude 3.7 + Web Search & & Anthropic & \\
    DeepSeek + Search & & DeepSeek & \\
    \bottomrule
  \end{tabular}
  \vspace{0.5cm}
  \caption{LLMs and Web Research Tools we evaluated. Each LLM was evaluated using a ReAct agent \cite{yao2023reactsynergizingreasoningacting} in both Live and RetroSearch mode. Web Research Products were evaluated by humans using their UIs, using the live web.}
  \label{tab:models-and-tools}
\end{table}

\subsubsection{Agent Architecture}

Deep Research Bench aims to be agnostic to agent architecture, permitting any agent following an action-observation loop with tool usage. For this initial establishment of the benchmark, we use a variety of models with the standard ReAct (Reason + Act) agent architecture \cite{yao2023reactsynergizingreasoningacting}. The approach follows a simple loop of thought, action, and observation to iteratively solve a task. To keep agent runs bounded, we apply an iteration budget of 50 actions per task, which is rarely exceeded. Before running the ReAct agent, we let an LLM select tips from a manually curated list and add it to the agent's task prompt. This is to prevent failures like agents trying to call/email someone or giving up to quickly.

When applying the ReAct architecture to newer-generation "reasoning" models, we found that it would often breakdown due to the explicit thought stage, asking a model to "generate a thought" based on the history of thoughts, actions, and observations to guide the choice of the next action. Reasoning models often reject prompts of this form, which we believe is due to guarding against users trying to jailbreak the model's internal thought process.

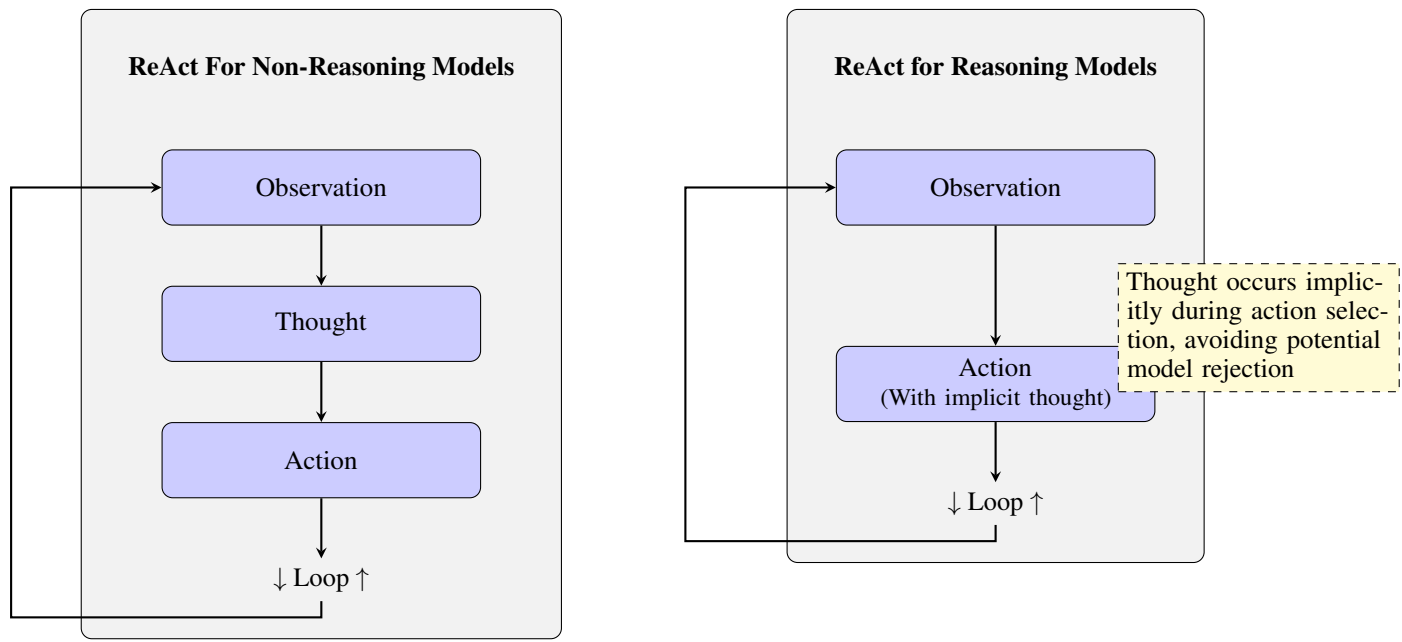
\begin{figure}[ht]
  \centering
  \begin{tikzpicture}[
      node distance=0.8cm and 4cm, 
      block/.style={rectangle, draw, fill=blue!20, text width=4cm, rounded corners, minimum height=1cm, text centered},
      arrow/.style={thick,->,>=stealth},
      title/.style={font=\bfseries}
  ]
  \node[title] (title1) {ReAct For Non-Reasoning Models};
  \node[block, below=of title1] (observe1) {Observation};
  \node[block, below=of observe1] (thought) {Thought};
  \node[block, below=of thought] (action1) {Action};
  \node[below=of action1] (loop1) {$\downarrow$ Loop $\uparrow$};
  \draw[arrow] (observe1) -- (thought);
  \draw[arrow] (thought) -- (action1);
  \draw[arrow] (action1) -- (loop1);
  \draw[arrow] (loop1) -- ++(0,-0.5) -| ([xshift=-2cm]observe1.west) |- (observe1);
  \node[title, right=of title1] (title2) {ReAct for Reasoning Models};
  \node[block, below=of title2] (observe2) {Observation};
  \node[block, below=1.6cm of observe2] (action2) {Action\\{\footnotesize (With implicit thought)}}; 
  \node[below=of action2] (loop2) {$\downarrow$ Loop $\uparrow$};
  \draw[arrow] (observe2) -- (action2);
  \draw[arrow] (action2) -- (loop2);
  \draw[arrow] (loop2) -- ++(0,-0.5) -| ([xshift=-2cm]observe2.west) |- (observe2);
  \node[draw, dashed, text width=3.5cm, fill=yellow!20, below right=0.5cm and -0.5cm of observe2] {Thought occurs implicitly during action selection, avoiding potential model rejection};
  \begin{scope}[on background layer]
      \node[fit=(title1)(loop1), draw, rounded corners, fill=gray!10, inner sep=0.5cm] {};
      \node[fit=(title2)(loop2), draw, rounded corners, fill=gray!10, inner sep=0.5cm] {};
  \end{scope}
  \end{tikzpicture}
  \caption{Comparison of traditional ReAct approach and the modified approach for reasoning models.}
  \label{fig:react-comparison}
\end{figure}

To work around this, we skip the explicit thought step for reasoning models, and instead rely on this thought occurring implicitly when we select an action. For reasoning models which do accept prompts asking for explicit thought, and therefore allow comparison of these approaches, we found that the implicit thought modification does not degrade performance. The difference between the flow for these agent approaches is represented in Figure \ref{fig:react-comparison}.

As a control sample, we also run each task instance with a single LLM prompt with no tools. This gives reference for how well each model performs outside of an agent framework and independently of any external additional input from reading the open web. Effectively, this measurement shows whether a model can remember or hallucinate its way to the correct solution. Results for this control are provided in the Appendix \ref{sec:memorization-analysis}.
\subsubsection{Tools for Agent Actions}

We provide each agent with two tools for web research: Google Search and Query Document.
Google Search utilizes the Serper API \cite{serperAPI}. The agent specifies a search query, and the tool returns a list of URLs and snippets in the same format as a call to the Serper API. We provide the agent with detailed instructions on how to Google effectively, including descriptions of useful features such as the \texttt{site:} filter operation, double quotes, and boolean operators.

Query Document gives an agent the facility to read a webpage and query relevant text content. An agent specifies a URL, typically taken from the results of a Google Search call, and a query to match to relevant excerpts. Excerpting limits the amount of context this tool produces. We maintain a very large excerpt size (65536 characters) to ensure we provide as much context as possible. In practice, most pages are returned from the tool without excerpting.

To read pages, we use multiple methods in sequence. We use Playwright \cite{playwright}, backed by a Firefox browser, to make a first attempt to extract the page, and follow this up with a direct HTTP request for the page. We use Playwright first because it handles slow and dynamic pages better than a simple direct request. If these approaches fail, we subsequently use the ScraperAPI service \cite{scraperAPI}. ScraperAPI facilitates access to hard-to-scrape pages, representing a reasonable best-effort attempt that realistic agents may make to obtain webpage content. In Section \ref{sec:retrosearch}, we discuss an alternative approach we use that's backed by frozen web page content.

\subsection{Evaluation Runs}
\label{sec:running-the-evaluation}

We ran each web research tool once per instance. We ran all other approaches (ReAct agents and toolless single prompts) twice per instance.
In the event that an attempt failed due to an invalid response format, we retried at least ten times. If none of the retries succeeded, we assigned the attempt a score of zero. 

All the results were collected over a period of two months to minimize divergence between the RetroSearch database and the live web.

For all LLM completions, we used a temperature of 0. When using Claude 3.7 Sonnet's extended thinking mode, we set the thinking token budget to 2048 tokens. For Gemini 2.5 Flash and Pro, we did not specify the thinking configuration, meaning thinking budget was determined automatically by the API \cite{geminiThinking}. For o3, we used the default reasoning effort of "medium" \cite{openaiReasoningEffort}.

\subsection{RetroSearch}
\label{sec:retrosearch}

\begin{figure}[h]
  \centering
  \includegraphics[width=0.99\textwidth]{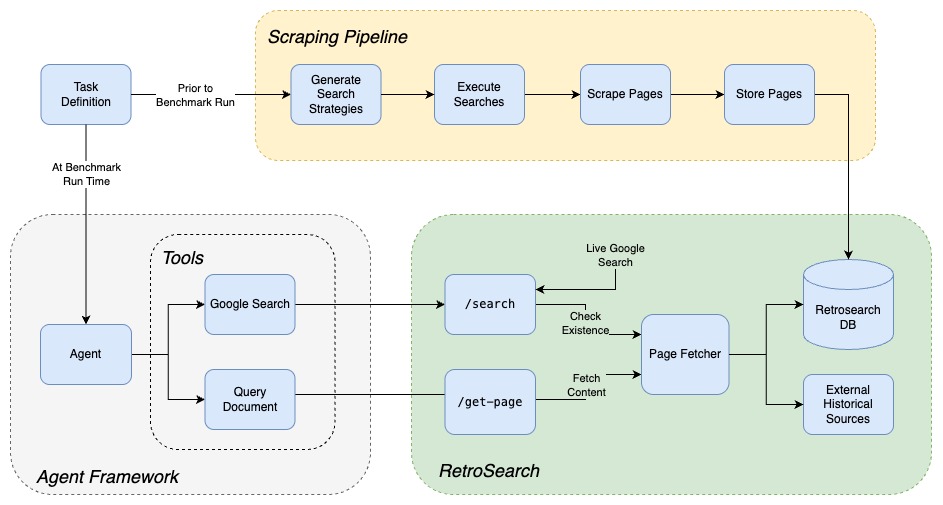}
  \caption{System architecture of Deep Research Bench using RetroSearch. This shows the flow from task definition through the scraping pipeline that populates the RetroSearch database prior to running the benchmark, and then how agents use RetroSearch via an API at the time of task evaluation.}
  \label{fig:retrosearch-arch}
\end{figure}

This section describes the architecture of the RetroSearch system and how our test agents interact with it. The high-level representation can be seen in Figure \ref{fig:retrosearch-arch}.

\subsubsection{Running a RetroSearch Query}

RetroSearch aims to emulate Google search (specifically, the Serper search API \cite{serperAPI}) as closely as possible, so as to minimize differences between live and "retro" agent runs. A single RetroSearch search query follows the following steps:
\begin{itemize}
  \item Run a live Serper search for the query
  \item Look up pages obtained from live search in the RetroSearch database and other archive sources
  \item If the page is not found in the RetroSearch database, remove it from the results
  \item Write new snippets from a sample of page content using a simple LLM
  \item Return the results in the original format of the Google results
\end{itemize}

This approach ensures a search experience for agents that is consistent with real search, but backed exclusively by pages we have a frozen candidate for. We use Gemini 2.5 Flash to write snippets of page content in all cases, using a sample of the first 5096 characters of cleaned Markdown content. We note that there is an opportunity for information leakage to occur here, as it is reasonable to expect that the ordering of live search results, due to its correlation with real-world relevance at the time of search execution, might subtly influence the agent's behavior. We do not take any steps to prevent this under the assumption that this effect is minimal, and intend to analyze this further in the future.

In addition to our RetroSearch database, we use the Common Crawl \cite{commoncrawl} as a fallback. Common Crawl is a large corpus of web pages that is updated regularly, with snapshots available for public access through its API. We found that this provides some useful pages, but is often incomplete on its own due to missing key sources that block CCBot, Common Crawl's web crawling bot. Particularly, most news sources are blocked, which can complicate researching current-events topics.

\subsubsection{Populating the RetroSearch Database}
\label{sec:retrosearch-populating}

We populate the RetroSearch database using crawling routines that we execute upon creating benchmark tasks, ultimately producing a single, large snapshot of pages per task. Table \ref{tab:task-type-instances} gives some average stats for the number of pages procured for each task type. For runs of the benchmark, we do not constrain each task to its specific snapshot, but instead allow each task to reference pages from any snapshot within the defined time window.

The crawling routine has two distinct phases: the crawling phase, and the scraping phase. The crawling phase involves a recursive query generation and execution routine: we first search the space of background information for the task subject, and then utilize this to generate specific search strategies which generate yet further searches. For example, for the task "List the most notable cybersecurity incidents between 2015 and 2024", after first reading relevant background information, we expand our search by following general strategies such as "search for major ransomware attacks", "search for high profile data breaches", "search for annual cybersecurity reports", and "search for financial damage estimations", and expand each of these recursively as we would the top-level approach. We deliberately avoid using agentic approaches to this so to reduce the risk of systematic biases that may result from similarity of the approach under test and the approach used to create the test dataset. We also expect that producing a large recursive dataset like this will produce a larger and more varied representation of the inherent noise of websearch, which is crucial for construct validity.

Once we assemble the results of all searches, we execute the scraping phase. Our scraping is exclusively handled with ScraperAPI \cite{scraperAPI}, as this allows us to scrape conveniently at scale and with substantial improvements over direct scraping and rudimentary proxying methods. This ensures that we maximize the number of pages we successfully scraped. We scrape raw HTML from pages, and store both the raw and processed data in our database. Typically, we serve our agents processed data in a Markdown format, which we keep consistent between live and retro agent runs. We use Trafilatura \cite{barbaresiTrafilaturaWebScraping2021} to convert HTML to Markdown.

The size of a typical crawl varies depending on the task type, as some tasks fundamentally require a larger number of pages. For example, some instances of the Find Number task may be answered by a single search and single page read if sufficiently well targeted, whereas a Gather Evidence task inherently requires searching and reading through significantly more pages.

\subsection{Web Research Products}
\label{sec:methods-web-research-products}

Finally, we compare LLM agents to a set of LLM-based web research products (see Table \ref{tab:models-and-tools}).

Importantly, these products could not be connected to our RetroSearch environment. So they use the live web, which may have changed in the time after the RetroSearch page crawling and scraping. Furthermore, we could not run the ~25\% of instances which depend on blocking access to certain web pages that contain answers worked out.

We gave the same prompt to the Web Research tools that we also used for the agents. OpenAI Deep Research always asks the user a clarification questions, which we answered with "You decide." Similarly, Gemini Deep Research presents the user with a research plan and asks for confirmation to start the research. We confirmed this without any modifications or clarifications to the original research plan. In instances where ChatGPT presented us with two different answers to choose from, we always picked the right one.

Consequently, we used a second prompt asking for the output in the same JSON format we're using for the ReAct agents (and instructing it not to do any further work or research). When asking for the formatting, we turned off "web search" where possible to prevent the web tool from conducting more research. At times, this two-stage setup posed challenges such as slight inconsistencies between the answer given by the web research tool originally, and the the answer provided after formatting. For example, the LLM might slightly mess up the formatting, forget to include a URL, or report an answer slightly different from the original answer.

We manually checked for inconsistencies, where feasible, and corrected them. We opted for this two-prompt approach because it mimics the behavior of the ReAct agents, where we ask it to solve the task first and then have a final formatting prompt at the end. We also saw instances where web research tools changed their behavior when given the task and formatting instructions in a single prompt and started hallucinating an answer without any research. Gemini Deep Research/Gemini Pro 2.5 via the web interface failed to follow our formatting instructions entirely. We therefore exported the research result to Google Docs, downloaded it as markdown, and then asked Gemini 2.5 Pro via the API for a correctly formatted solution based on that.

\subsection{Evaluating Performance}
\label{sec:methods-evaluating-performance}

\subsubsection{Scoring Metrics}

We evaluated tasks using various metrics, depending on the type of the task. Table \ref{tab:task-scoring-methods} gives an overview of the scoring methods used for each task. For Derive Number, Find Number, and Find Original Source, we assign a binary score (0 or 1) depending on whether all requirements are met. We aim to overcome the limitation of not assigning partial credit by creating a sufficiently high number of task instances to provide a nuanced picture of agent performance.

\begin{table}[h]
  \centering
  \renewcommand{\arraystretch}{1.2}
  \begin{tabular}{l>{\raggedright\arraybackslash}p{3.5cm}>{\raggedright\arraybackslash}p{6.5cm}}
  \toprule
  \textbf{Task} & \textbf{Scoring Method} & \textbf{Success Criteria} \\
  \midrule
  Compile Dataset & Precision, Recall, F1 & Comparison of the rows of the dataset returned by the agent with the ground truth \\
  Derive Number & 0/1 Binary Score & Number is correct (or for some tasks, number is within a reasonable range) \\
  Find Dataset & Recall & Proportion of URLs in the list of required dataset(s) found \\
  Find Number & 0/1 Binary Score & 1. Number correct, \newline 2. Number backed up by excerpt from URL, \newline 3. Source at least as reliable as ground truth source \\
  Find Original Source & 0/1 Binary Score & URL is in a list of permissible truth URLs \\
  Gather Evidence & Recall & Comparison with a minimal list of evidence items to be found \\
  Populate Reference Class & Precision, Recall, F1 \newline (sometimes only Recall) & Comparison of agent list with ground truth list \\
  Validate Claim & Absolute Difference in Assigned Probability & Comparison of agent assessment with human researcher assessment \\
  \bottomrule
  \end{tabular}
  \caption{Scoring methods for different types of tasks}
  \label{tab:task-scoring-methods}
\end{table}

For Compile Dataset, Find Dataset, Gather Evidence, and Populate Reference Class, the agent solution comprises multiple items.

All Find Dataset and Gather Evidence instances are scored using recall, defined as
\begin{equation}
  \text{recall} = \frac{\text{number of from the ground truth list found}}{\text{number of items in the ground truth list}}.
\end{equation}

To avoid rewarding agents for generating an excessive number of items, we usually also impose an instance-specific maximum for the number of items that can be returned.
The ground truth list is chosen such that any reasonable answer with this maximum number of items will contain at least the items therein. For example, a gather evidence task asking about the ongoing projects of a company might allow 15 items to be returned, and require that the top 3 most prominent projects (according to our subjective judgement) are included.

For Compile Dataset and Populate Reference Class, it is sometimes possible to exhaustively compile a correct list, and so while for some instances we use recall scoring, for others we use the F1 score, defined as
\begin{equation}
  \text{F1 score} = 2 \cdot \frac{\text{precision} \cdot \text{recall}}{\text{precision} + \text{recall}},
\end{equation}
where
\begin{equation}
  \text{precision} = \frac{\text{number of correct items}}{\text{total number of items returned}}.
\end{equation}

For Validate Claim, we use the absolute difference between the agent's answer and the answer given by human analysts. The difference is normalized such that the possible scores range from 0 to 1.

\begin{equation}
  \text{VC score} = 1 - \frac{|\text{agent answer} - \text{human answer}|}{\text{max}(\text{human answer}, 1 - \text{human answer})}.
\end{equation}

This score is admittedly subjective since it is based on the assessment of human analysts. For example, it is possible that a superhuman agent would receive a lower score than a worse agent, if it is able to provide a better assessment than the human analysts. This is a necessary complication that arises from including tasks with ambiguous or unknown ground truth.

For some of the tasks, we employ LLMs to assist in the evaluation process. One reason is to overcome the limitations of string matching, for example when to equate "MIT" with "Massachusetts Institute of Technology." Another reason is to assess whether a source provided by the agent is at least as reliable as the ground truth source.

\subsubsection{Aggregating Scores}

After grouping by approach and task, we calculate the mean score to measure each approach's performance on each task. If Approach A achieves a higher mean score than Approach B on the Find Number task, that is evidence that Approach A is better at the requisite skills.

However, it is invalid to compare scores between different tasks. If Approach A achieves a higher score on Find Number than Derive Number, that does \textit{not} imply that the approach is better at finding numbers than deriving them. The task instances are too disparate for such comparisons to be meaningful.

\subsection{Evaluating Traces}
\label{sec:trace-evaluation}

Evaluating traces is a non-trivial task. Average traces have between 5 to 15 agent loop iterations. This corresponds to 10 to 30 steps for thinking agents with a Observation - Action loop, and 15 to 45 LLM decisions for non-thinking agents with an Observation - Thought - Action loop, see Figure \ref{fig:react-comparison}. Each decision has some history, and often contains a web payload, sometimes a lengthy one.

Across hundreds of such traces, exhaustive manual inspection is infeasible. After reading through many traces ourselves, we came up with a taxonomy of tractable failure modes for \emph{thoughts}, \emph{actions}, and \emph{observations}---see Figure \ref{tab:failure-modes}.

\begin{table}[H]
  \centering
  \begin{tabular}{p{2cm}p{3cm}p{11cm}}
  \toprule
  &\textbf{Failure Mode} & \textbf{Description and Example} \\
  \midrule
  \textbf{Thought checks} & & \\
  & Hallucination & Hallucinations in the thought. \vspace{0.1em} \newline \textit{The agent observed that a paper cites Schwalb (2024) and thinks that it should look for \texttt{Schwalb, Karl (2024)}.} \\
  \addlinespace
  & Included tool call & A thought includes a tool call. \vspace{0.1em} \newline \textit{The agent includes a tool call (using proper JSON syntax) in its thought.} \\
  \addlinespace
  & Mistake in reasoning & A thought makes incorrect inferences. \vspace{0.1em} \newline \textit{The agent finds results for a benchmark in the original paper and assumes that it can infer the SotA number from this, failing to account for the possibility that these numbers may be outdated.} \\
  \addlinespace
  & Reasoning for invalidated approach & A thought advocates for an approach that was already tried without success. \vspace{0.1em} \newline \textit{The agent reasons that it should try accessing OpenAI's website, despite the fact that it has already tried this unsuccessfully, since they block bots.} \\
  \addlinespace
  & Gullibility & A thought uncritically takes information from low-quality sources at face value. \vspace{0.1em} \newline \textit{The agent believes a number from a social media post or some low-quality market research report that was likely generated automatically.} \\
  \addlinespace
  & Missed the point & A thought misses the point of the task. \vspace{0.1em} \newline \textit{Tasked with looking up the amount of rainfall in a certain region according to a source, the agent thinks it should look up how this source collects its data.} \\
  \addlinespace
  \textbf{Action checks} & & \\
  & Hallucination & Hallucinations in the tool call. \vspace{0.1em} \newline \textit{The tool call (e.g. the actual Google search query) hallucinates information that is not present in the thought or the trace so far.} \\
  \addlinespace
  & Incompatible tool call & The tool call is compatible with the preceding thought. \vspace{0.1em} \newline \textit{The last thought suggests to explore the page further (e.g. scrolling down), but the tool call is for using Google to look for other pages altogether.} \\
  \addlinespace
  & Mistake in tool usage & The tool call is used incorrectly.\\
  \addlinespace
  & Incorrect tool call & The tool call is incorrect. \vspace{0.1em} \newline \textit{A tool that accepts a URL and a single question (to return the paragraphs most likely containing the answer) is called with multiple questions.} \\
  \addlinespace
  & Repeated tool calls & Used a tool call it already made before. \vspace{0.1em} \newline \textit{The agent repeats the same Google search query multiple times.} \\
  \addlinespace
  & Forgets information & The tool call could be better with information from the trace so far. \vspace{0.1em} \newline \textit{The agent already saw the link to an article, but then uses Google to look for the same article again.} \\
  \addlinespace
  \textbf{Observation checks} & & \\
  & Blocked & The agent was blocked from accessing a website. \\
  \addlinespace
  & Garbled text & The agent got served garbled text as response to its request of a page.\\
  \addlinespace
  & Page not found & The agent got a 404 error. \\
  \bottomrule
  \end{tabular}
  \caption{Taxonomy of failure modes for thoughts, actions, and observations}
  \label{tab:failure-modes}
\end{table}

There are several caveats:
\begin{itemize}
  \item These failure modes are neither supposed to be mutually exclusive, nor are they supposed to be exhaustive.
  \item Their classification is not always clear-cut and their automated classification is not always correct. So we restrict our attention to those of interest that we can classify correctly with high confidence: Action checks for hallucinations, repeated tool calls, and forgetting information.
  \item Some failure modes don't apply to a given step (the observation checks, for example, are only applicable to tools that access web pages, but not to, say, Google) or a given architecture (thinking models, for example, don't generate\footnote{For those thinking models that do generate thoughts, we did not log them.} accessible thoughts that we can analyze).
\end{itemize}

To improve the accuracy of the automated classification of our taxonomy of failures, we found that
\begin{itemize}
  \item o3 was the best model (outperforming Gemini 2.5 Pro and Sonnet 3.7 non-thinking models) to perform these checks,
  \item the trace had to be presented suitably, i.e. with the right amount of context and with annotations highlighting the part to pay attention to,
  \item making one call to o3 per check-step pair (e.g. we make one call for checking for hallucinations in the second \emph{thought} of a given trace) achieves a much higher accuracy than simply presenting the entire trace and asking for a list of issues, but this means evaluating traces is more expensive and limited than running the original agents,
  \item asking LLMs more open-ended questions like ``List any issues you see with the following trace'' has both a high false positive rate and a low true positive rate--in particular, this only catches a small fraction of the failures that we would like to catch.
\end{itemize}

\section{Results}
\label{sec:results}

\subsection{Quantitative Results}
\begin{figure}[H]
  \centering
  \includegraphics[width=0.95\textwidth]{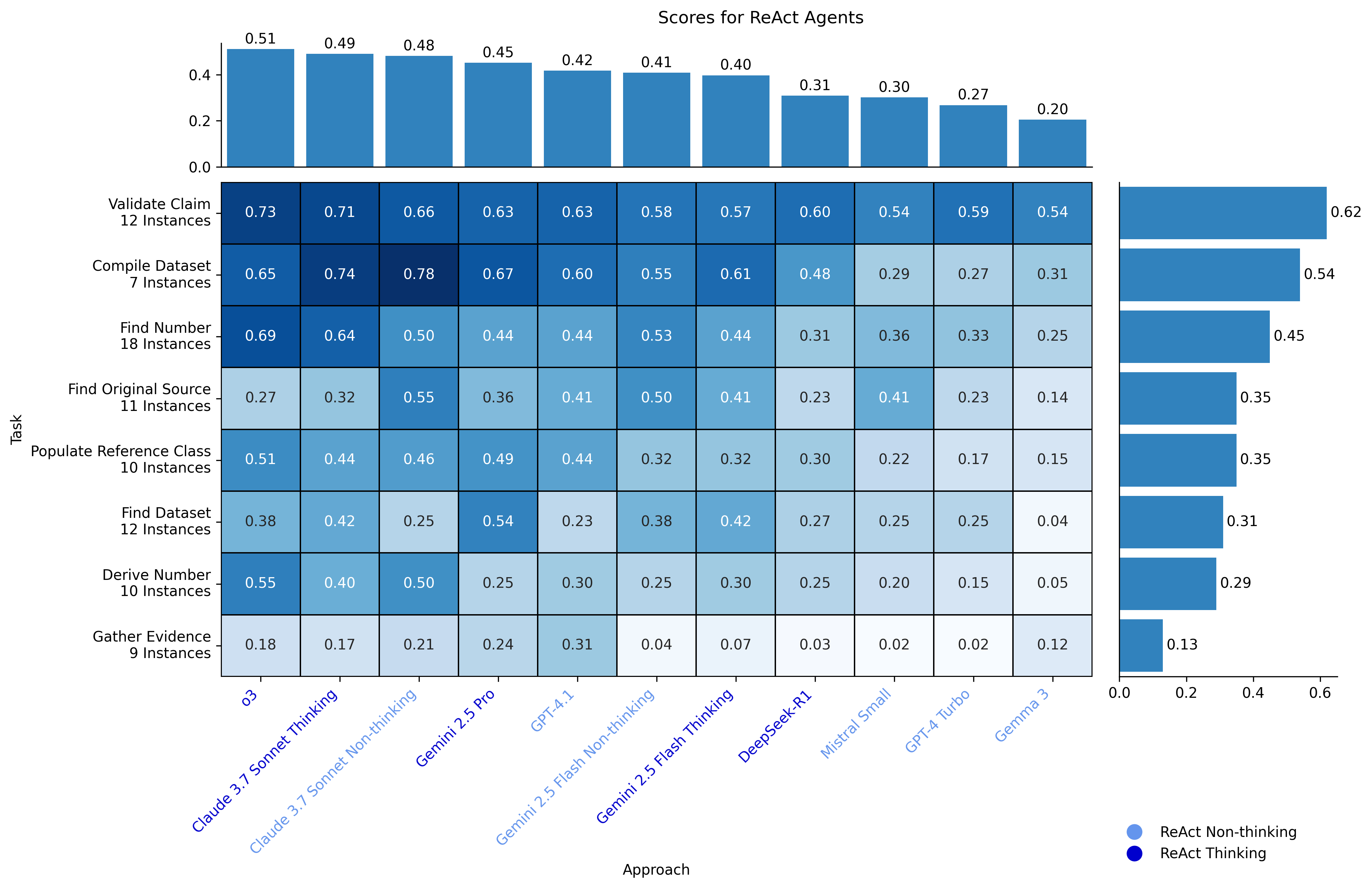}
  \caption{Scores across tasks and LLMs on the full set of 89 instances. Thinking models (which use the implicit-thought ReAct architecture) are colored in dark blue and non-thinking models (which use the regular explicit-thought ReAct architecture) in light blue.}
  \label{fig:quantitative-results-agents}
\end{figure}

\begin{figure}[h]
  \centering
  \includegraphics[width=0.99\textwidth]{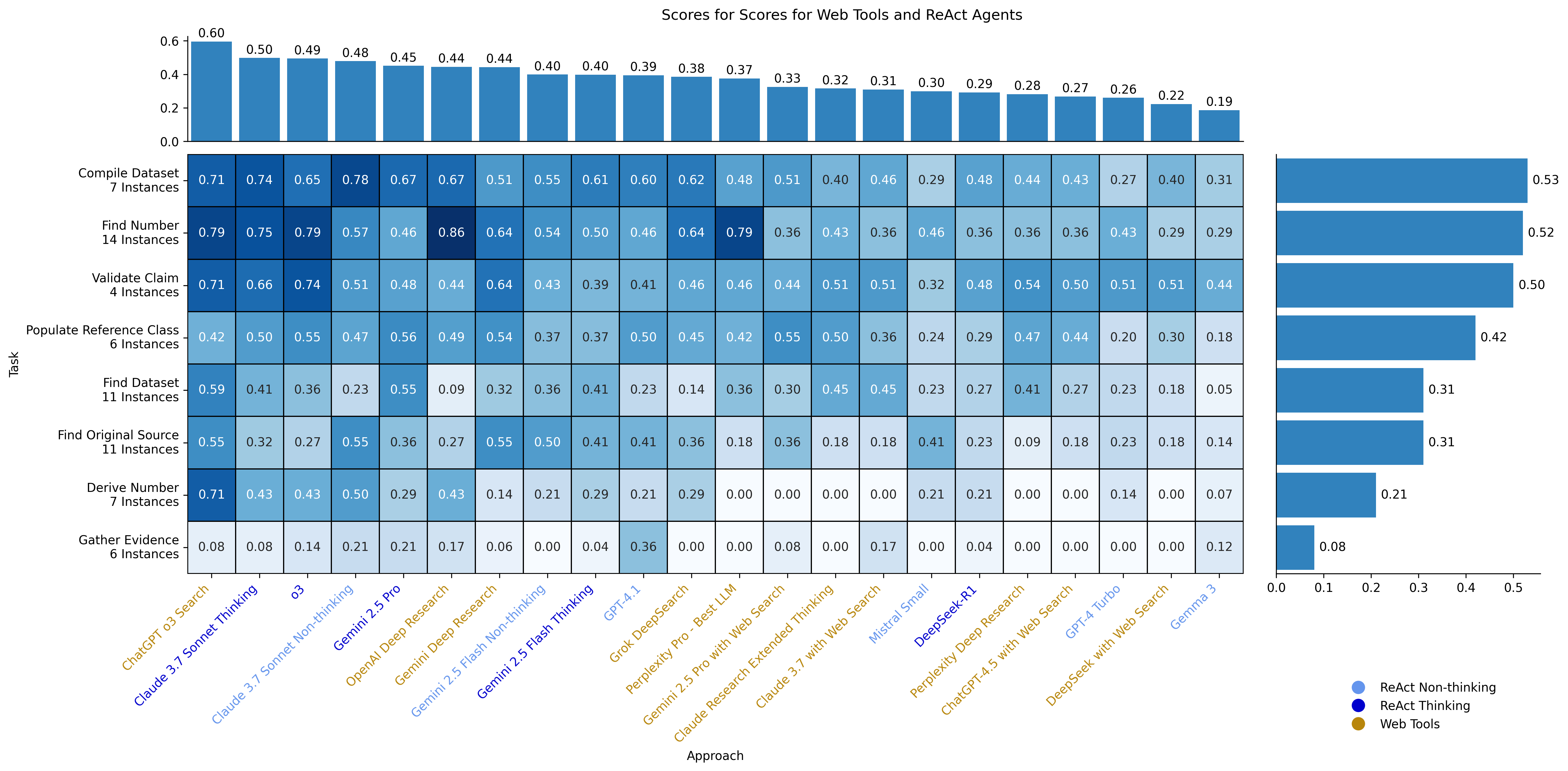}
  \caption{Scores for LLM agents and commercial web research products on the set of 66 instances that exclude those relying on blocking URLs from being accessed.}
  \label{fig:quantitative-results-web-vs-agents}
\end{figure}

Our main results are for web research agents using LLMs over their APIs, in Figure \ref{fig:quantitative-results-agents}. See Figure \ref{fig:heatmap-scores-webtools} for evaluations of web research agents as they appear in commercial "research", "search", "deep research", "deep search" tools, and Figure \ref{fig:quantitative-results-web-vs-agents} for a combined view. The Appendix has other charts more directly comparing models, such as by frontier lab or open source models.

Several observations speak to the basic fidelity of Deep Research Bench to real-world usage of web research agents. We see clear, significant progress since the release of GPT-4-Turbo. We see that the closed-weight frontier is substantially beyond the open-weight frontier. And, we see that larger models (Gemini 2.5 Pro) significantly outperform their smaller counterparts (Gemini 2.5 Flash).

\subsubsection{Frontier Performance}
The highest performance we observe is 0.51, from an o3 agent. Our rough sense is that the noise ceiling of Deep Research Bench --- the expected score for a ``perfect'' agent, different from 1 due to subjectivity, potential errors, and the use of LLMs in scoring functions --- is currently somewhere around 0.8. Our instances are such that a smart generalist researcher given ample time would reach this ceiling --- so, we can conclude that frontier agents under low elicitation substantially underperform smart generalist researchers who are given ample time.

Very crudely, we can use GPT-4-Turbo's score of 0.27 to say that about 45\% of the gap between smart generalist researchers and frontier agents as of April 2024 (this is the Turbo checkpoint we used) was closed one year later, by April 2025. Our subjective impression from manual inspection of agent traces (see Section \ref{sec:qualitative-assessment}) is that this exaggerates the picture, but not by much.

It is also notable that DeepSeek-R1 is comparable, and very likely better, as a driver of a web research agent than GPT-4-Turbo. While larger open models may have achieved this earlier, this puts a rough upper bound on the gap between the best closed model and the best open model for usage in web research agents at 8 months.

\subsubsection{The Big Three}
Frontier models from Anthropic, Google, and OpenAI are roughly tied; in our domain, no lab has a clear advantage. This contrasts interestingly with other research benchmarks (Humanity's Last Exam \cite{phanHumanitysLastExam2025}, for example, where o3 dominates).
Our suspicion is that Deep Research Bench tests systematic planning, thoroughness, and reliability to a greater degree, while Humanity's Last Exam focuses more on crystallized knowledge and deductive reasoning.

\subsubsection{RL-on-CoT}
One might wonder about the extent to which the recent RL-on-CoT paradigm represents a step change, as opposed to incremental progress. Our results weakly suggest incremental progress for the deep research domain; while the latest frontier non-thinking model, GPT-4.1, does underperform the recent clutch of thinking models, the gap is much less pronounced than the gap between the open-weight and closed-weight frontier.

We see that models trained with RL-on-CoT with thinking disabled perform very similarly under explicit-thought ReAct prompting to agents with thinking enabled and without explicit-thought prompting.

This is evidence that explicit-thought prompting with thinking disabled elicits CoTs of similar quality to those generated in thinking mode.


\subsection{RetroSearch}

The above results in the "live" environment most closely match web research agents in the real world. But as the internet is changing, going forward the results from the RetroSearch environment will be more reliable and should be considered canonical, as we report on our public-facing website. Here we investigate the faithfulness of the RetroSearch environment to these tasks on the live internet.

\begin{figure}[h]
  \centering
  \includegraphics[width=0.6\textwidth]{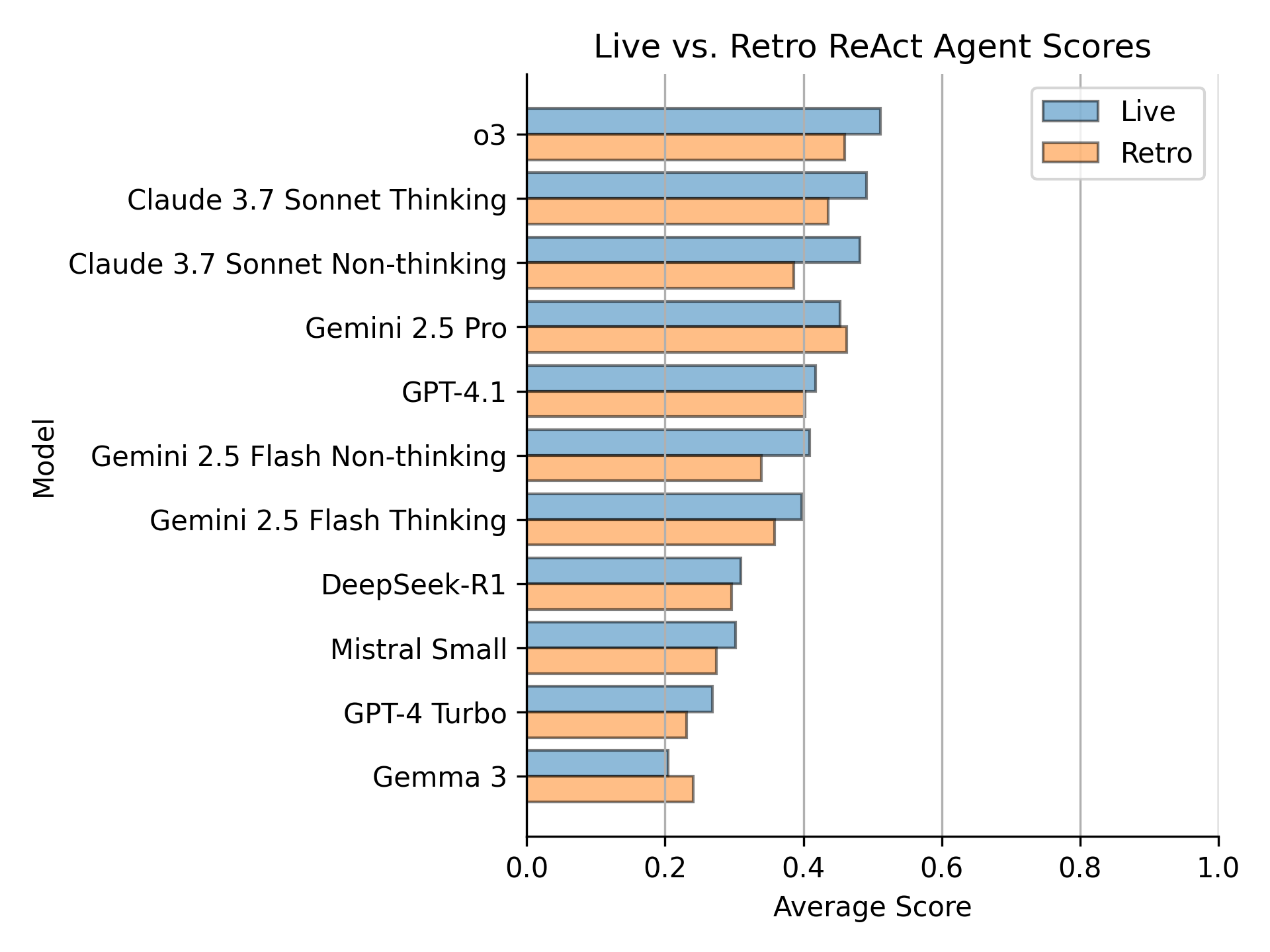}
  \caption{Average scores for Live and Retro variants of the ReAct agents for each LLM}
  \label{fig:live-retro-average-scores-by-approach}
\end{figure}

Table \ref{tab:live-retro-sorted-approaches} compares the performance of ReAct agents with access to the live web to those with access to the RetroSearch database. The order of the LLMs is broadly similar, with o3, Claude 3.7 Sonnet and Gemini 2.5 Pro among the top for both Live and Retro agents. Mistral Small 3, GPT-4 Turbo and Gemma 3 are at the bottom. However, the average scores and the order of the LLMs are not identical. These results are also shown in Figure \ref{fig:live-retro-average-scores-by-approach}.

\begin{table}[h]
  \centering
  \renewcommand{\arraystretch}{1.2}
  \begin{tabular}{ll|ll}
    \hline
    \multicolumn{2}{c|}{\textbf{Live}} & \multicolumn{2}{c}{\textbf{Retro}} \\
    \textbf{LLM} & \textbf{Average Score} & \textbf{LLM} & \textbf{Average Score} \\[4pt]
    \hline
    \rule{0pt}{14pt}
    o3 & 0.51 & Gemini 2.5 Pro & 0.46 \\
    Claude 3.7 Sonnet Thinking & 0.49 & o3 & 0.46 \\
    Claude 3.7 Sonnet Non-thinking & 0.48 & Claude 3.7 Sonnet Thinking & 0.44 \\
    Gemini 2.5 Pro & 0.45 & GPT-4.1 & 0.40 \\
    GPT-4.1 & 0.42 & Claude 3.7 Sonnet Non-thinking & 0.39 \\
    Gemini 2.5 Flash Non-thinking & 0.41 & Gemini 2.5 Flash Thinking & 0.36 \\
    Gemini 2.5 Flash Thinking & 0.40 & Gemini 2.5 Flash Non-thinking & 0.34 \\
    DeepSeek-R1 & 0.31 & DeepSeek-R1 & 0.30 \\
    Mistral Small & 0.30 & Mistral Small & 0.27 \\
    GPT-4 Turbo & 0.27 & Gemma 3 & 0.24 \\
    Gemma 3 & 0.20 & GPT-4 Turbo & 0.23 \\[4pt]
    \hline
  \end{tabular}
  \caption{Average scores for Live and Retro variants of the ReAct agents}
  \label{tab:live-retro-sorted-approaches}
\end{table}

\label{sec:results-retrosearch}

\begin{figure}[h]
  \centering
  \includegraphics[width=0.8\textwidth]{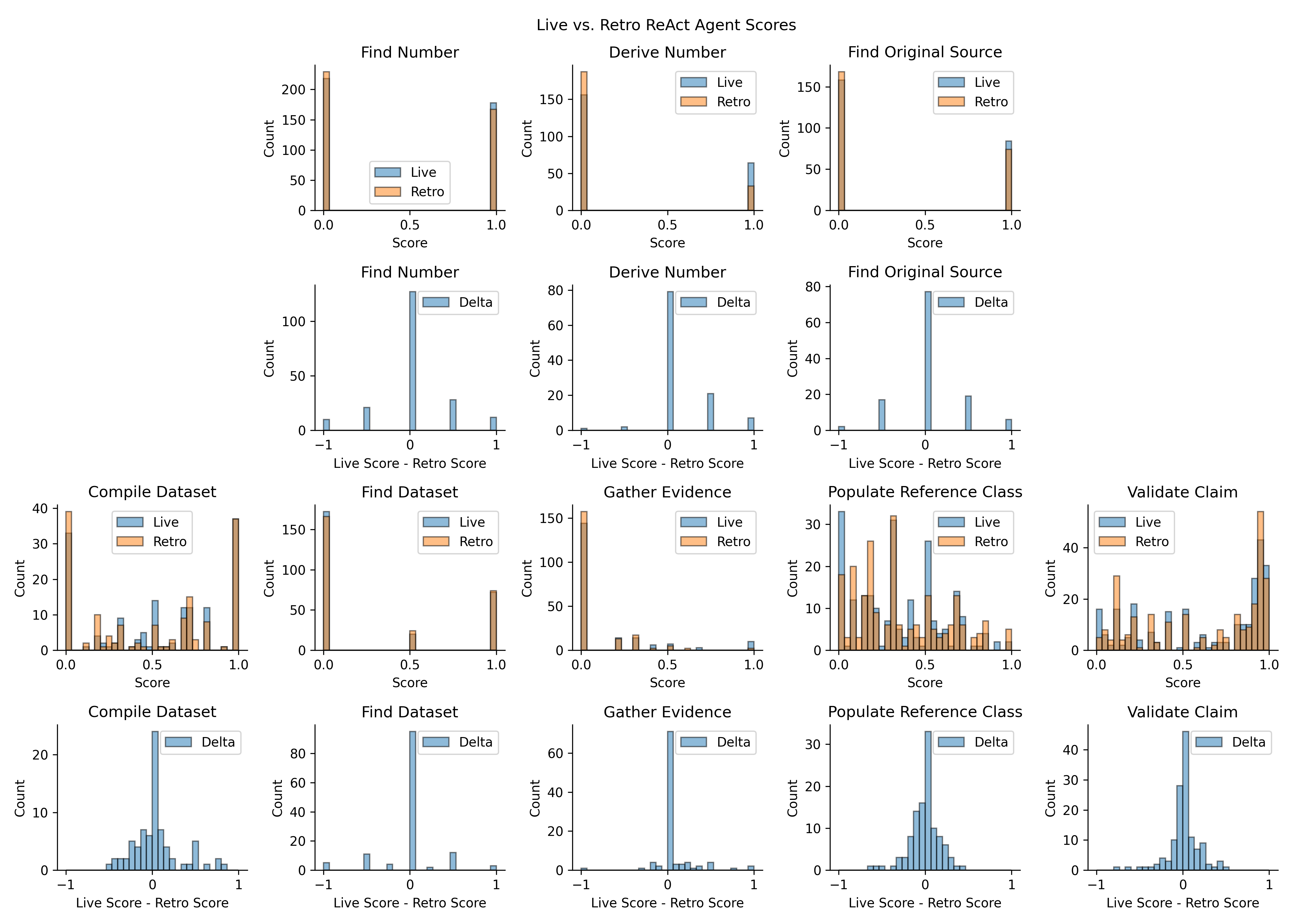}
  \caption{Scores for Live and Retro variants of the ReAct agents. For each task, the upper histogram shows the score distribution across all instances and LLMs. The lower histogram shows the difference between the Live and Retro scores. We calculate the mean score across all repeats of an instance and LLM before taking the difference. This makes the difference insensitive to the order of the scores.}
  \label{fig:live-retro-histograms}
\end{figure}

Figure \ref{fig:live-retro-histograms} shows a more detailed breakdown of the performance of Live and Retro agents. If Live and Retro agents performed identically, the score distributions would be the same and the score deltas would all be zero. This is not the case. Some of the discrepancy is due to random variation in LLM calls: an agent may happen to run different search query or reason differently, even when the same information is available to it.

A potential source of systematic divergence between live and Retro runs is a lack of relevant information in the RetroSearch database due to flaws in the crawling and scraping phases. Another possible factor contributing to differences between Live and Retro scores is divergence between the RetroSearch database and the live web as pages are added, removed and updated. This could provide an advantage to Live agents if new information is published. But it could also benefit Retro agents if information is taken down. However, we expect such divergence to be minimal since the results were collected over a relatively short timeframe of two months.

\subsection{Runtime and Accuracy}

\begin{figure}[H]
  \centering
  \includegraphics[width=0.8\textwidth]{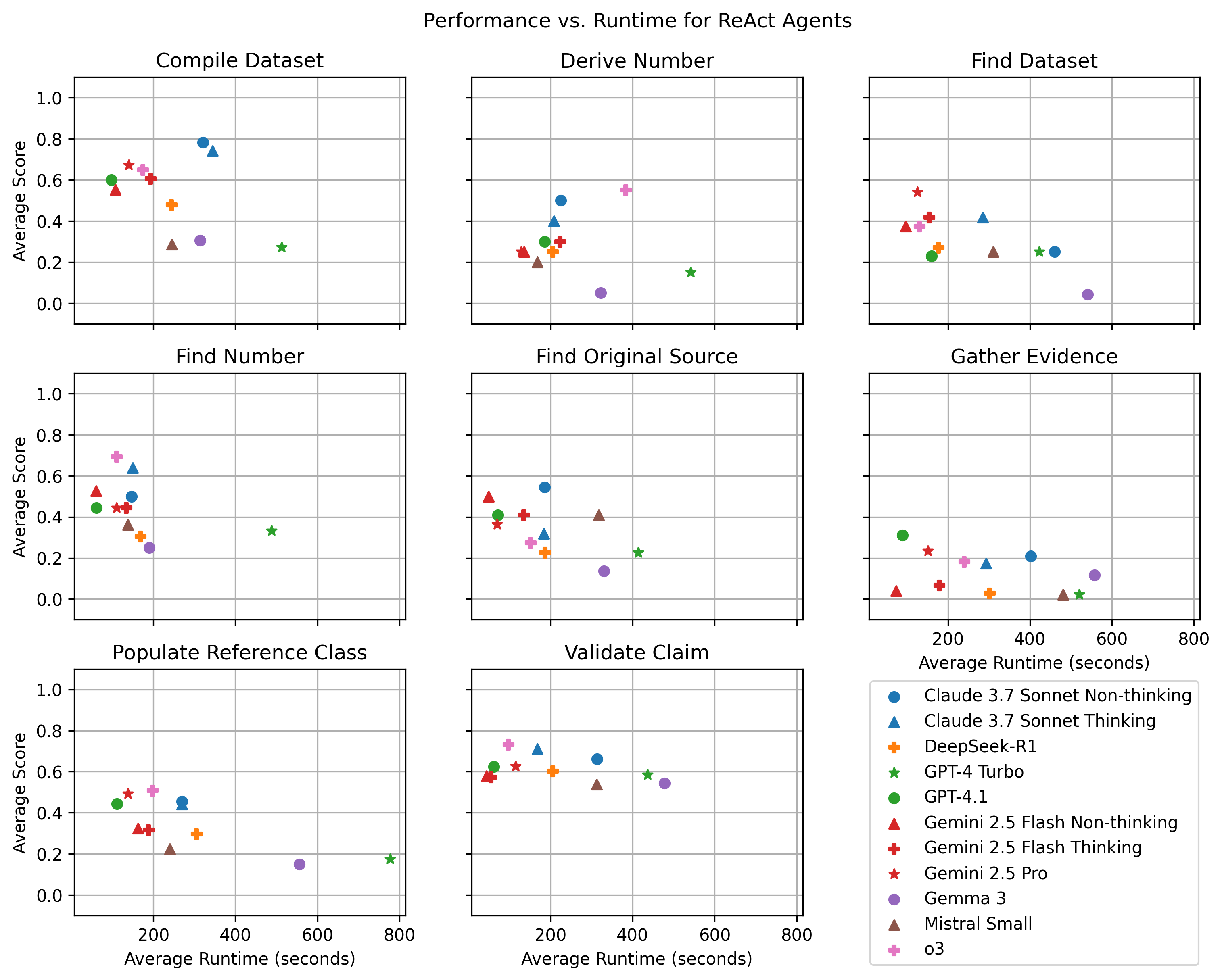}
  \caption{Performance and runtime for ReAct agents. Runtimes generally varied from about 1 minute to 10 minutes.}
  \label{fig:scores-vs-runtime}
\end{figure}

Figue \ref{fig:scores-vs-runtime} shows the average runtime of agents against their scores on each task. In the Section \ref{sec:prompting} we described that agents sometimes halted very quickly even when they clearly did not have a sufficiently correct or thorough answer.

Longer runtime was not generally associated with a higher score. For example, o3 and gemini-2.5-pro agents were generally faster than average and scored near the top in most tasks.

\subsection{Commercial Web Research Products}

\begin{figure}[H]
  \centering
  \includegraphics[width=0.99\textwidth]{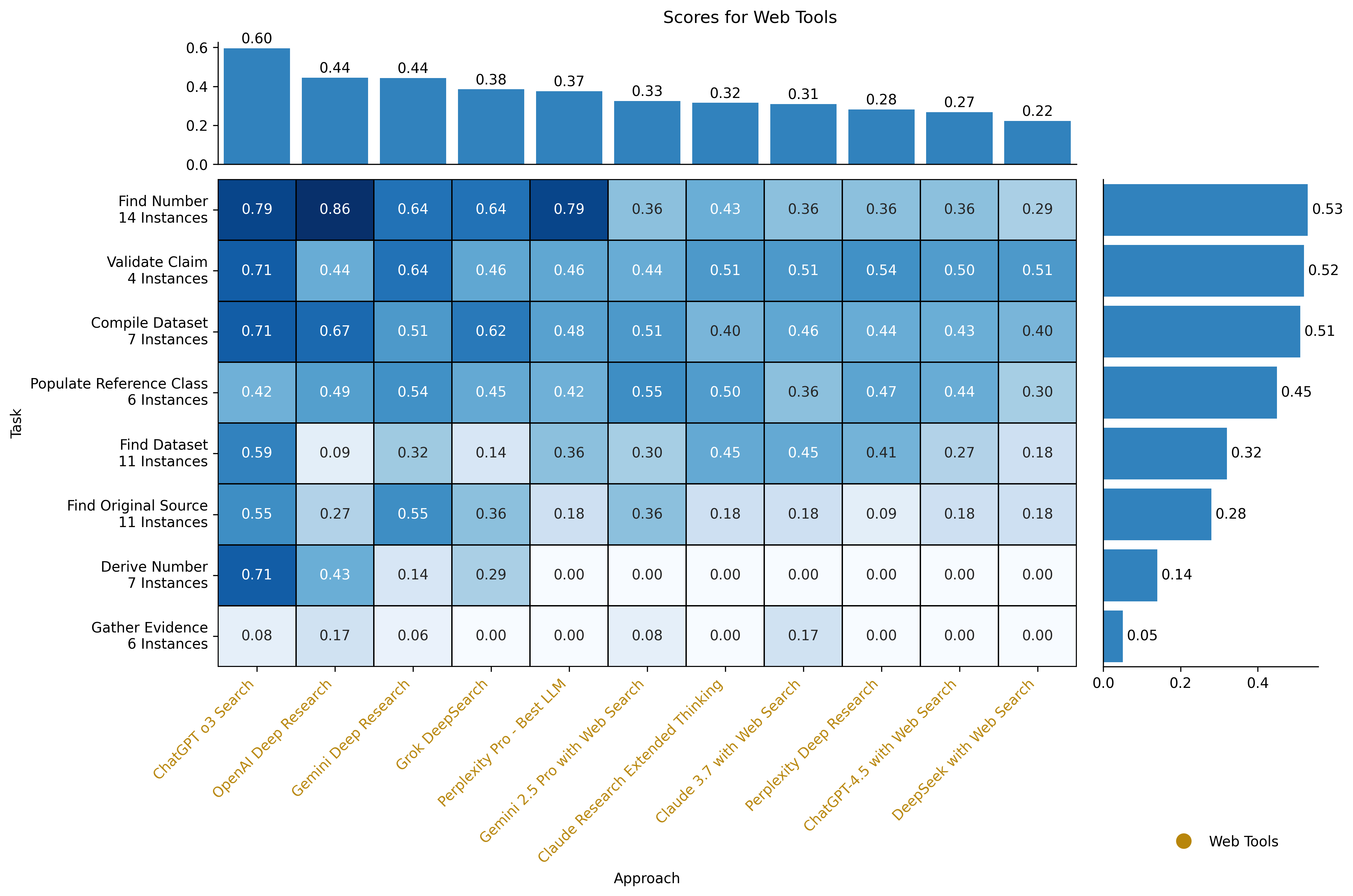}
  \caption{Scores for web research tools}
  \label{fig:heatmap-scores-webtools}
\end{figure}

Our analysis of commercial tools (Figure \ref{fig:heatmap-scores-webtools}) shows ChatGPT-o3 as the clearly best at these tasks, beating out even OpenAI's o3-driven Deep Research. One clear qualitative pattern was ChatGPT-o3's propensity to validate its answers before completing, which may explain why it had high scores across all tasks (except Gather Evidence for which no agent under any conditions scored highly). Other commercial web research products had at least one style of question where they performed consistently poorly, comparable to the quality of a web agent of an older, smaller model.

This suggests that, when it comes to research with simple, fairly clear success criteria, returns to the deep-research post training that OpenAI carried out to produce Deep Research are limited. To the extent that users express a preference for Deep Research, then, this may be due to a preference for more background information rather than accuracy or genuine comprehensiveness, for example in cases where people want a much longer treatment of a subject they are unfamiliar with.

In a similar vein, we observe that the standard Perplexity Pro workflow outperforms their Deep Research offering.

However, we do not find this for Gemini: Gemini Deep Research performed noticeably better than Gemini-2.5-pro with Web Search. Claude Research Extended Thinking also performed slightly better than Claude-3.7 with Web Search.

Although our sample size for evaluating commercial tools is small, it's worth noting the superior performance of ChatGPT-o3 to our o3 agent. While we are not confident that this result will hold in a larger sample, subjectively we did observe increased thoroughness from the web-UI model, which we speculate may be caused by a system prompt encouraging such behavior. It also seems likely that OpenAI have trained o3 to work with their internal search tool.

\subsection{Automated Trace Evaluation}
\subsubsection{Failure Rates}
  As detailed in Section \ref{sec:trace-evaluation}, due to cost restrictions, we ran only a subset of full automated trace evaluations. We ran evaluations on six LLM models (Deepseek R1, Gemini 2.5 Flash, GPT-4 Turbo, Mistral 3.1 Small, Claude 3.7 Sonnet, Claude 3.7 Sonnet Non-Thinking), across 16 to 32 traces per agent. The results are shown in Table \ref{tab:failure-rates}. Results for failure modes other than those listed here were not deemed reliable enough to include.

\begin{table}[h]
  \centering
  \begin{tabular}{l|ccc|c}
  \toprule
  \textbf{Model} & \multicolumn{3}{c|}{\textbf{Action Failure Rates} (per step)} & \textbf{Steps} (per run) \\
  \cmidrule(lr){2-4} \cmidrule(lr){5-5}
                        & Hallucination & Repeated Tool Calls & Forgets Information & Total \\
  \midrule
  DeepSeek R1           & 0.159 & 0.044 & 0.115 & 4.1 \\
  Gemini 2.5 Flash NT   & 0.011 & 0.198 & 0.090 & 10.5 \\
  GPT-4 Turbo           & 0.019 & 0.293 & 0.356 & 13.1 \\
  Mistral 3.1 Small     & 0.039 & 0.107 & 0.162 & 10.1 \\
  Claude 3.7 Sonnet     & 0.014 & 0.236 & 0.111 & 10.8 \\
  Claude 3.7 Sonnet NT  & 0.014 & 0.175 & 0.171 & 14.4 \\
  \bottomrule
  \end{tabular}
  \caption{Failure rates across different models and failure modes. NT = Non-Thinking variant.}
  \label{tab:failure-rates}
\end{table}

Table \ref{tab:failure-rates} supports our subjective impressions from manual inspection of traces:
\begin{itemize}
  \item Hallucinations have not decreased much. GPT-4 Turbo's hallucination rate is 0.019, which is within statistical noise of Sonnet 3.7's 0.014. DeepSeek R1's is very high (0.159). But this does not seem to be the central issue for models' performance.
  \item Repeatedly calling the same tool with the same or similar inputs (or infinite looping, in the extreme case) remains an issue, but seems to occur less frequently with newer models that perform better on Deep Research Bench.
  \item Models got better at dealing with longer context windows, as evidenced by the lower rates of forgetting information in newer and better models.
  \item Comparing Claude-3.7-Sonnet Thinking vs Non-Thinking, thinking models are more likely to get stuck repeatedly calling the same tools without adapting to failures, while non-thinking models are more likely to forget important information.
\end{itemize}

\subsubsection{Predicting Task Scores}
Using these three failure modes as predictors in a regression model to predict task scores, we find that forgetting information is the strongest predictor. Specifically, we find that a linear regression model with an intercept of -2.142 shows that forgetting information has the largest negative coefficient (-0.843, p=0.014), followed by hallucinations (-0.653, p=0.055) and repeated tool calls (-0.165, p=0.626). However, the model's predictive power is quite limited, with an R-squared of 0.093 on the training set and 0.129 on the test set, suggesting these failure modes alone do not explain most of the variation in task performance.

Note that this does not allow us to conclude anything causal. This would require rerunning agents from a partial trace ending in a \emph{corrected} mode to see how much this improves performance. We leave this to future work.

\subsubsection{Predicting Failure Rates}
One might wonder whether models tend to hallucinate more as traces get longer. To test this, we compute the correlation\footnote{Concretely, we compute for each position in the trace the fraction of traces that have a hallucination at that position. This way we get a $[0,1]$-valued vector, a function of the step index. We then compute the correlation coefficient between this vector and the step index.} of the step index with the fraction of hallucinations at each step across all traces for Sonnet 3.7. We find that the correlation is positive (0.48), but not statistically significant (p=0.076).

\subsection{Qualitative Assessment}\label{sec:qualitative-assessment}
In this section, we give a qualitative account of the performance of frontier agents along the dimensions that influence performance on Deep Research Bench most strongly, based on manual inspection of agent traces. For each dimension, we attempt to give a concrete picture of capabilities and limitations via representative examples.

\subsubsection{Planning, Strategy}
Agents are capable of spontaneously formulating somewhat systematic approaches to search tasks, but their strategies fall short of those that an intelligent and motivated human would generate, given ample time to solve the task.

As a representative example, when tasked with compiling a list of facilities meeting certain criteria, we observed an o3 agent first issuing a general search, then systematically going through each candidate facility shown in the results and checking whether it fits the criteria or not. This is a reasonable approach, but a more robust strategy would have been to enumerate the firms that could plausibly own such facilities, and conduct a thorough and targeted search for each firm, focusing on sources that commonly report on such things.

Moreover, this capability manifests rather unreliably; difficult search problems are often attacked in much less sophisticated ways. As an example, when tasked with finding the best-performing instance of a certain technology, we observed Gemini 2.5 Pro issuing a few google searches that went directly after the answer, stumbling across a leaderboard, and immediately reporting the (incorrect) top number. This is a far cry from the strategy a competent researcher would use.

\subsubsection{Tactics, Adapting to New Information}
Agents show some ability to use the information they find to adapt their research tactics dynamically, but can fail to respond to what they observe as intelligently as they might --- again, a smart and motivated human would do better.

One of the more impressive examples of adapting to information on the fly that we observed involved Gemini 2.5 Pro recognizing a pattern in URLs that it had to trawl to compute a sum, and then systematically querying them one by one. Although this is something a human would easily figure out, it represents a significant improvement on GPT-4-Turbo's attempt, where Turbo found a single number in the sum and immediately reported it as the final answer.

As an example of limited ability on this dimension, when tasked with another instance of finding the highest-performing technology in some category (similar to the task mentioned in the previous subsection), we observed an o3 agent repeatedly ignoring search results mentioning something that a human, after a little thought, would have realized is a promising candidate to investigate.

\subsubsection{Thoroughness, Optimization}
Under simple prompting (see Section \ref{sec:prompting}), on tasks where a range of solutions of varying quality are possible, models show a tendency towards satisficement rather than optimization. Models are prone to adopting minimum-effort approaches, and will often conclude that they have completed the task at the earliest opportunity (see also Section \ref{sec:reasoning-gullibility}). This behavior is especially pronounced in o3 and DeepSeek-R1.
Roughly, our assessment is that in about

For example, we observed o3 immediately stating that the data necessary to answer a query it had just received was not available, and returning a guess without performing any research.
And, an commonly taken by frontier agents to finding relevant datasets was to keywordize the query --- occasionally including a known data provider in the search --- and return results from the first few pages.

For tasks that have a single solution, we observed that thoroughness behaviors such as cross-referencing multiple sources, sanity/consistency checking, highly systematic search, etc, do not occur spontaneously.

\subsubsection{Reasoning, Gullibility}\label{sec:reasoning-gullibility}
While Deep Research Bench does not rely heavily on reasoning writ large, here we highlight a specific reasoning error that is responsible for a sizable portion of failures (perhaps 5-10\% in the top-performing models): When deciding whether something they've found represents a correct solution to a task, agents are prone to false positives, making incorrect inferences that a moderately intelligent human never would.

As an example, we observed a Sonnet 3.7 agent concluding that a certain document was the original source of a claim, but a little thought about the organization that produced the document shows that this is effectively impossible. And, when tasked with finding a statistic where multiple versions are available from different sources, and one source is clearly of much higher quality than the rest, we observed o3 immediately accepting a poor-quality version, even though the high-quality source was the second hit for its search, and the task explicitly states that the highest-quality number available must be reported.

Although this failure mode substantially affects the performance of frontier models here, our impression is that it is notably less common than in older or smaller models.

\subsubsection{Reliability, Hallucinations}\label{sec:reliability-hallucinations}
Egregious non-humanlike failures are still common. As one of many such examples, we've observed Gemini 2.5 Pro confusing an economic statistic it found in a document for the statistic requested in the task instance, where the required statistic was also in the document, and the mistake was obvious at a cursory glance. And we've seen o3 repeatedly googling a fake identifier for an non-existent government document that it believed contained a requested number.

\subsubsection{Googling Skills}
Models show mixed understanding of and ability to use Google. More often than not their queries are reasonable, but it's uncommon to see a search that uses reasoning about which terms are likely to be used in which kinds of document, and intelligently uses site filters and other operators. Search terms tend to be overly influenced by the instance's phrasing, and we occasionally observe outright mistakes. For example, all frontier models have been observed to wrap very verbose queries in exact-match quotes, predictably leading to an empty result. And, when searching for multiple entities, each of which should be expected to appear in a separate document and not together, models will sometimes include all target entities in a single query, without using the OR operator.

\subsubsection{Summary}
Overall, our qualitative assessment is that, under low elicitation, frontier agents fall short of intelligent humans trying hard and given ample time on difficult open-web research tasks. However, tasks that require $\sim$20 minutes of research time for a human unfamiliar with the domain, that do not require much systematic planning, and where opportunities to erroneously conclude that a correct answer has been arrived at are few, are solved by frontier AI, albeit with less-than-perfect reliability.

\section{Discussion}
\label{sec:discussion}

In this paper, we introduced Deep Research Bench, a novel benchmark for evaluating the capabilities of large language models (LLMs) on open-web research tasks. Deep Research Bench extends the landscape of existing benchmarks by providing a set of tasks that realistically reflect real-world open-web deep research tasks, and by providing a controlled environment, "RetroSearch", that allows us to evaluate the performance while controlling for the continually-changing web. We evaluated the performance of a set of eleven LLMs, finding that state of the art closed weight models can makes significant progress on easier tasks, but have not yet achieved human-level performance on the hardest tasks. Our results are available on a public leaderboard at \url{https://drb.futuresearch.ai/}. We provided evidence that offline "RetroSearch" agents perform comparably to "live web" agents in terms of relative rankings, suggesting the possibility to assess the performance of future LLMs using our controlled environment. In addition, we compared the performance of the eleven LLMs against a set of nine proprietary web research products. In terms of results, we present both quantitative and qualitative assessments of the performance of current state-of-the-art LLMs and common failure modes that still set them apart from competent human researchers.

Deep Research Bench is subject to a number of limitations. Our current approach limits interaction with web pages to reading and querying static content. A more capable agent would be able to interact with web pages, e.g. by clicking on links, scrolling, and providing a more dynamic and responsive interface, thus more directly mimicking human interaction with the web. This difference may become especially relevant as UI-interaction capabilities such as Claude's Computer Use develop. The fact that the current benchmark only comprises 89 task instances, plus the fact that we were only able to do two runs per agent, makes rigorous statistical analysis difficult. The requirement for some of the tasks to be graded with the help of LLMs introduces a further source of potential noise.

While using the RetroSearch approach offers advantages in being able to produce a consistent and frozen view of the web for agents, it also comes with downsides. First, this requires substantial crawling work to be carried out up-front, and these crawls must carry the representative data for any task instance, as well as the correct solution. This can lead to a number of systematic biases. For example we might miss key pages or fail to predict approaches an agent may take. If we miss a page containing key information, then this can cause the agent to consistently fail tasks. Relatedly, if we miss pages which might contain poor information, either because it's below the standard our eval expects, or because it's blatantly misleading, then this can make tasks easier. This latter point is especially important, as it's easy to predict the path to the right answer, and therefore ensure it's present, but not as easy to predict the many paths to wrong answers.

Our solution to this has been to try to cast a wide net spanning the space of reasonable searches, therefore maintaining as much of the complexity that a live agent would encounter while solving these tasks. We also potentially introduce bias by removing the inherent unpredictability associated with reading pages from the open web, such as failures due to bot-blockers or webpages occasionally being down, and replacing it with a well-behaved API and database under our control. RetroSearch does not protect us entirely from the march of time. As time progresses, so too do training cutoff dates, and newer LLMs will contain crystallized pre-training knowledge which exceeds that of older LLMs and invalidates our snapshot dates. As such, we would expect to recreate new snapshots and re-evaluate the benchmark entirely periodically.

Note, however, that this does not restrict us from using retro evaluations to compare new agent architectures as they are developed. Even if we deem that our RetroSearch database is sufficiently representative for current models and agents, it is not strictly clear that it will be so for future, and likely more capable, models and agents. We find it probable that a more capable agent would be more likely to follow more predictable paths towards the correct answer. However, this is not guaranteed and the RetroSearch approach may prevent us from observing novel and unforeseen approaches to solving tasks.

When running evaluations for the commercial open-web research products, we only ran a single iteration of each product and task instance due to time constraints. This increases noise in the results. Based on our subjective experience from past work with LLMs and similar web research products, we believe that results still accurately reflect real capabilities, but we cannot exclude effects from random variation in LLM outputs. As web research products are not optimized for producing output in the specific structured JSON format we require for our tasks, subtle errors could have been introduced at this point. For example, we observed instances where, for a Gather Evidence task, a web research product returned a list of 19 URLs, but only included 17 of them in the final formatted output. Sometimes, web research products returned an empty JSON, even though they found a solution. We checked outputs and corrected them manually, but we might have missed subtle errors. For Gemini Deep Research, we were unable to obtain any correct JSON output and had to produce the formatted output based on an exported markdown file created from the Gemini Deep Research output. This might have introduced additional errors. Lastly, web research products differed in the tools available to them. For example, Claude Web Search was not able to read PDFs, as far as we can tell.

In the future, we commit to updating and extending this benchmark continuously. We will add more task instances as well as task types. In addition, we will expand the number of LLMs and agent architectures we evaluate. For example, we would like to study inter-run variation  in more detail by running a large number of repeats for each instance and a given LLM. As discussed in Section \ref{sec:prompting}, our results represent performance under a low-elicitation regime. We plan to explore the performance of frontier agents under a high-elicitation regime in future work. In the future, we also intend to investigate cost and runtime as part of our evaluations.

\section*{Acknowledgments}
This work was supported by Open Philanthropy.

\printbibliography










\appendix

\section{Memorization Analysis}
\label{sec:memorization-analysis}

\begin{figure}[h]
  \centering
  \includegraphics[width=0.99\textwidth]{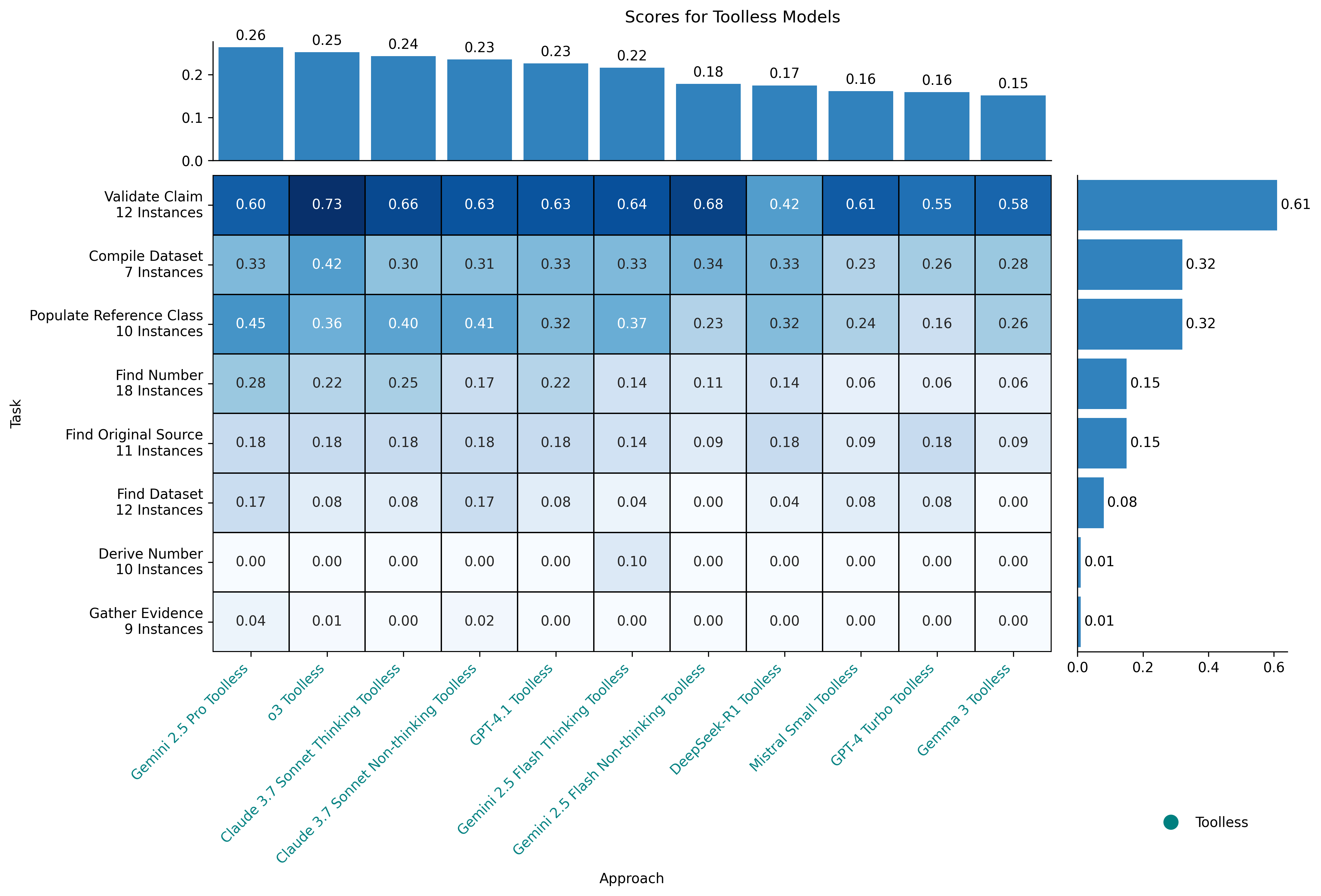}
  \caption{Scores across tasks and LLMs on the full set of 89 instances for Toolless agents. Toolless agents are single-prompt "agents" that execute the task definition a single task prompt and follow-up JSON formatting prompt with no tools, therefore relying purely on training data memory.}
  \label{fig:memorization-analysis}
\end{figure}

Each LLM has significant "memory" of information obtained from its training data. As such, it is in principle possible for agents to score highly on certain tasks purely based on recall from this memory. To analyze this, we conducted evals on our full set of instances using "Toolless" agents comprising a single task prompt and an additional follow-up JSON formatting prompt. No tools are available to these agents, therefore there is no source of external information. Any solution to tasks from these agents is purely based on the LLM's training data and any context that can be gleaned from the prompt, and from the LLM's training data.

Figure \ref{fig:memorization-analysis} shows the scores for each LLM on the full set of instances.
The most notable result is that for Validate Claim. For this task, Toolless agents perform comparably to Live and Retro ReAct agents, with an average score of 0.61 vs 0.62 for Live ReAct agents. We also observe the trend in capability from less-performant models to more-performant models, with o3, Claude 3.7 Sonnet and Gemini 2.5 Pro among the top for both Live and Retro agents and Mistral Small 3, GPT-4 Turbo and Gemma 3 at the bottom. This result suggests that external information has little effect on the capability to perform well on this task, but it is not yet clear to us whether this is due to internalized memory of the topics of tasks or due to an enhanced capability to judge claim validity based on intrinsic priors. Other than Validate Claim, the performance on other tasks exhibits less pronounced memorization effects, with Derive Number and Gather Evidence in particular being cases where Toolless agents are effectively incapable of remembering a correct solution. Curiously, we do not see a strong correlation between the performance of Toolless agents and training window cutoff, as exemplified by o3 consistently performing well despite its relatively old (December 2023) training window cutoff.

We intend to continue to perform analysis of this nature in order to inform us about the viability of our individual instances as time and models progress, and further explore the reasons for instances of enhanced performance in of Toolless agents.

\section{Additional Results}

\begin{figure}[H]
  \centering
  \includegraphics[width=0.6\textwidth]{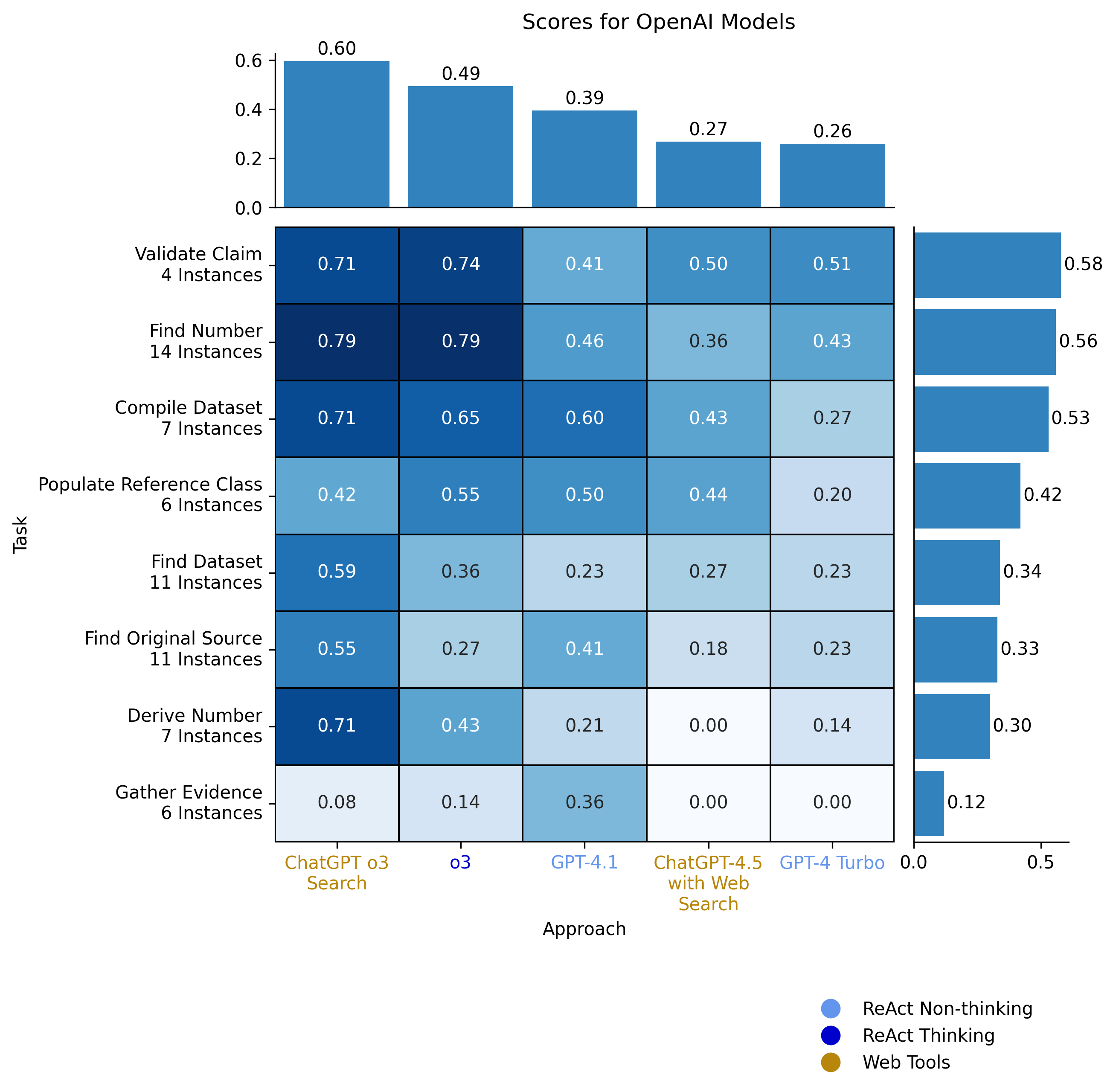}
  \caption{Scores for OpenAI models}
  \label{fig:heatmap-scores-openai-models}
\end{figure}

\begin{figure}[H]
  \centering
  \includegraphics[width=0.6\textwidth]{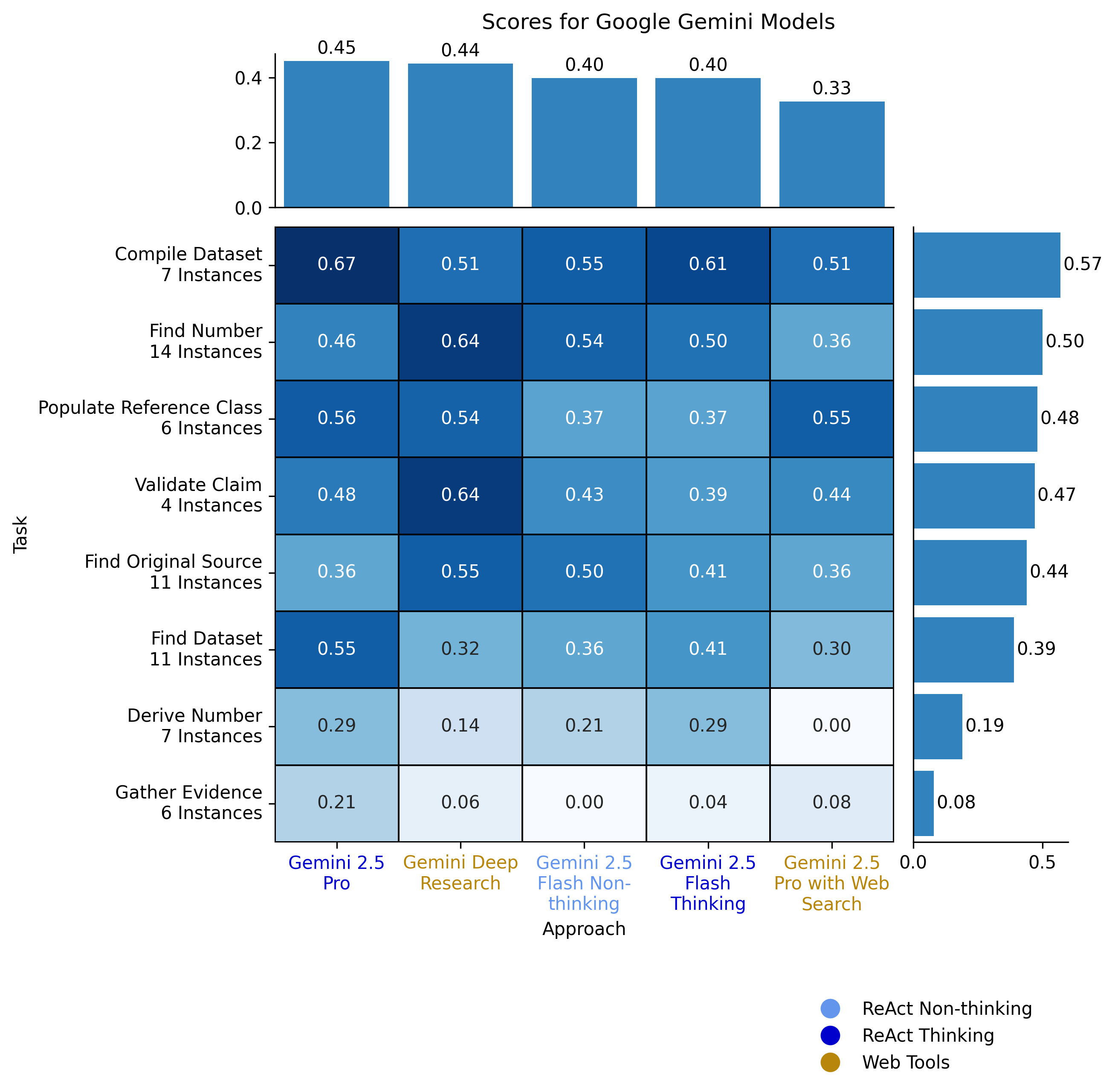}
  \caption{Scores for Google Gemini models}
  \label{fig:heatmap-scores-gemini-models}
\end{figure}

\begin{figure}[H]
  \centering
  \includegraphics[width=0.6\textwidth]{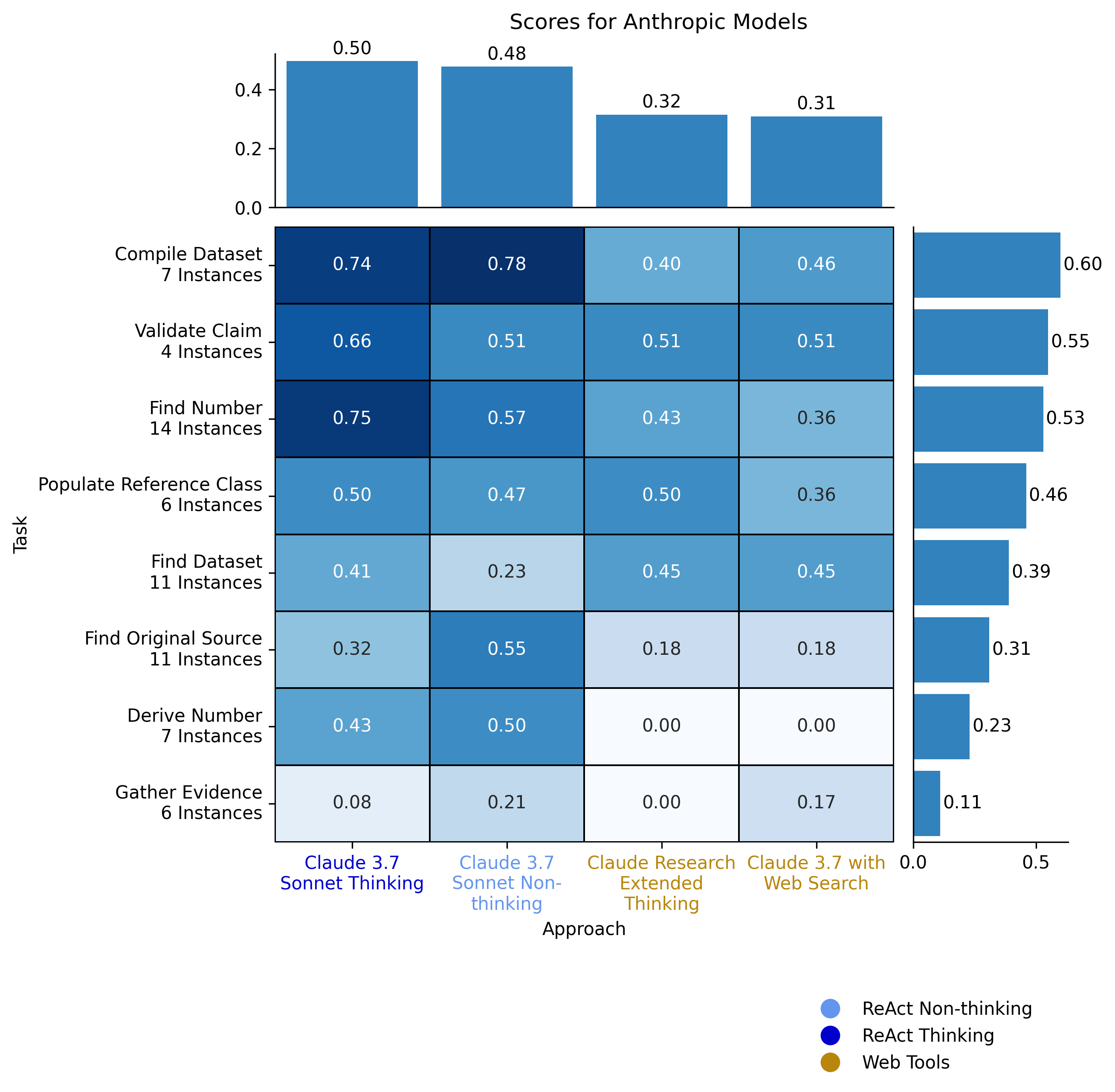}
  \caption{Scores for Anthropic models}
  \label{fig:heatmap-scores-anthropic-models}
\end{figure}

\begin{figure}[H]
  \centering
  \includegraphics[width=0.6\textwidth]{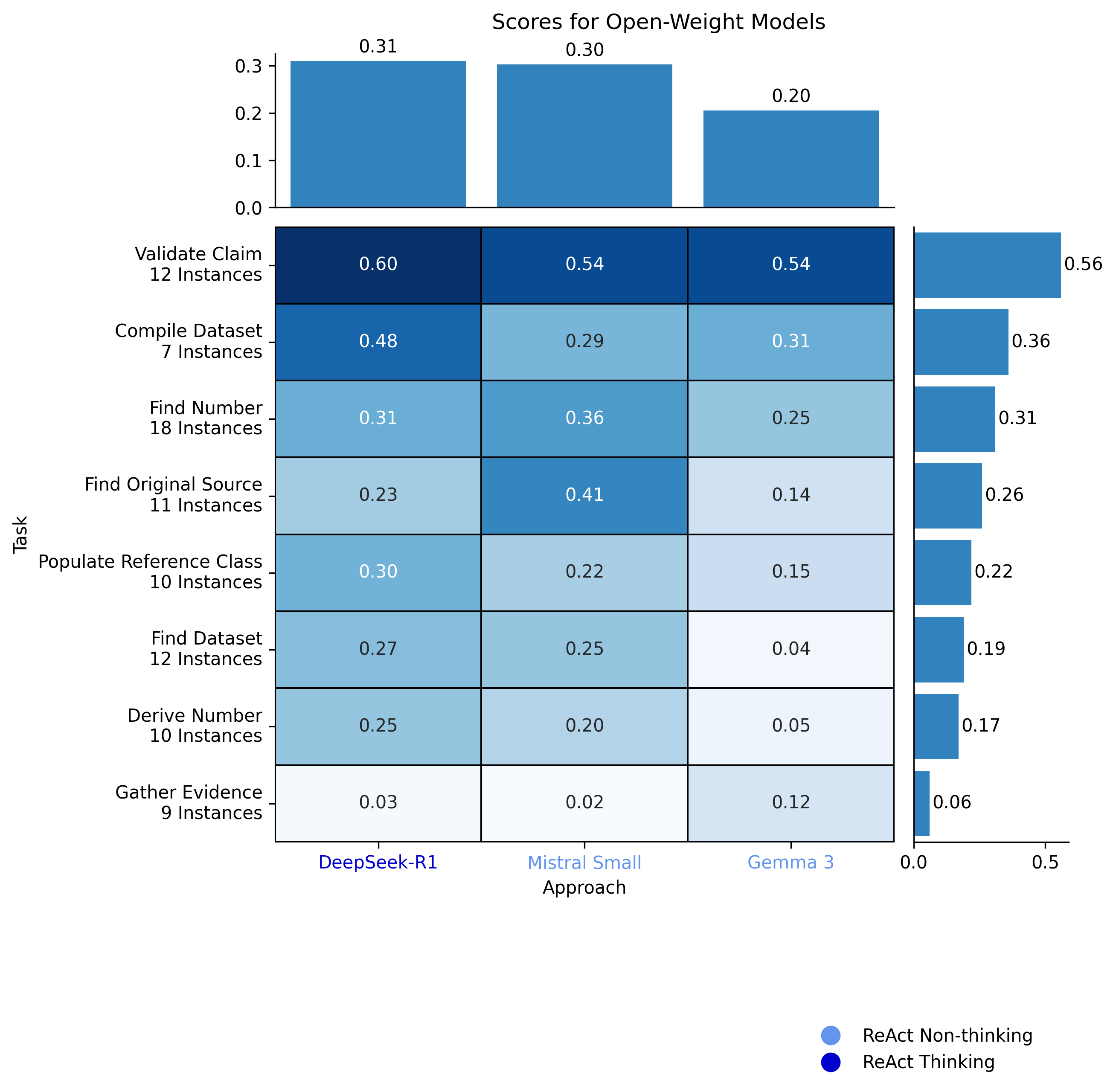}
  \caption{Scores for open weight models}
  \label{fig:heatmap-scores-open-weight-models}
\end{figure}

\section{Example Tasks}

\subsection{Find Number}

\exampletaskbox{Highest reported pass@1 agent performance on the Cybench benchmark}{Please find the value of the following number: highest reported agent performance on the Cybench benchmark; the most precise number you can find (i.e. to as many decimal places as possible). Your answer should have the following unit: average percentage of tasks solved on the first attempt. You should return the most reliable answer you can find.
If you find multiple sources and numbers, return the one that is the best (most reliable).
If you can't find a source referencing a specific number, set the success flag to False and give placeholders for the other fields.
Otherwise, set the success flag to True and return the source, value, unit, and a supporting snippet from the source.
Note that your task is to find a source stating, or directly implying, e.g. by a table, graph, or a statement in different units, the number I am after
Your task is \textit{not} to work this number out yourself by combining information from multiple sources and making deductions.
You must include a snippet from the source you get the number from, supporting your answer. The snippet MUST contain the number you've been asked for, and should be extensive enough that an external evaluator can tell that the source really does provide the number as you claim.}

\subsection{Find Dataset}

\exampletaskbox{Number of big AI labs in China}{
  Please find datasets that are relevant to my query: "How many big AI labs (say, having trained an AI model with over 1e24 FLOPs) are there in China?"
If you can't find any datasets, set the \`{}found\_datasets\`{} flag to False and return an empty list.
Otherwise, set the \`{}found\_datasets\`{} flag to True and return a list of links to the datasets with reasoning why you chose them and which columns are relevant to my query.
For example, if I ask for "datasets about GDP per capita in UK during the industrial revolution" and you found a dataset that has estimated UK GDP, UK population, and many other economic indicators from 1600 onwards, you should return the link to the dataset and your reasoning should note that the industrial revolution started around 1750 (so the time requirement is satisfied) and that GDP per capita is not a column, but it can be calculated as GDP/population (which are present in the dataset).

Note that while many datasets are csv or excel files, they can also be tables, charts, other visualizations on websites or in pdf files (often papers), or even just high quality summary statistics (like quartiles, mean, median, etc.) if everything else fails.
Wikipedia, for example, has many datasets in the form of tables; in this case, just return the link to the article, not the table (e.g. https://en.wikipedia.org/wiki/Nuclear\_reactor, not https://en.wikipedia.org/wiki/Nuclear\_reactor\#Table\_by\_date).
Papers often have datasets in the form of tables or charts; in this case, just return the link to the paper, not the table or chart.
Government websites often have datasets in various (inconvenient) formats.
If in doubt, prefer \\
- datasets in a csv file or a table over interactive visualizations which are harder to access. (Many interactive visualizations are accompanied by a csv download link; preferably return the csv link.),\\
- datasets, tables, or charts over just summary statistics, \\
- up-to-date datasets (i.e. when there is a newer dataset available, prefer the newer one),\\
- non-paywalled datasets over paywalled ones.\\

You should return at most 10 datasets.
}

\subsection{Find Original Source}

\exampletaskbox{AI Generated Propaganda}{
The source https://economictimes.indiatimes.com/tech/technology/ai-generated-propaganda-as-effective-as-original-propaganda-finds-new-study/articleshow/107978780.cms contains the following claim: "Artificial intelligence-generated propaganda is almost as effective and persuasive as real propaganda, a team of researchers have found after a study involving more than 8,000 US adults.".
What is the original source of this claim?
We say a source is original if its relevant information cannot be traced back to another URL that it links to or got its information from.
So, for example, a blog post referencing a recent Ukrainian advance would not be "original", but the Reuters article announcing it (and whatever website the blog author likely got it from - directly or indirectly) would be.
However, if Reuters discussed the US announcing something, the original source would likely point to an announcement on a US government website, not Reuters.
For published papers, return the DOI (https://doi.org/10.xxxx/xxxxx), if possible. (This may not be possible for preprints on e.g. arXiv.)
If you cannot find the original source, set success to false and solution to null, otherwise set success to true and solution to the original source.}

\subsection{Validate Claim}

\exampletaskbox{Tesla Model H}{Evaluate the likelihood of the following claim: "Tesla announced a to-be-released 'Model H' car that would run on hydrogen fuel cells, as opposed to battery power."

Secondary sources: https://www.ipmi.org/news/tesla-unveils-plan-first-hydrogen-powered-vehicle-model-h-2026}

\subsection{Derive Number}

\exampletaskbox{Time for first open model to beat PaLM 540B on MMLU}{Your task is to figure out the value of the following number:
\\
\\
The number of months between the release of PaLM 540B and the release of the first model to outperform it on MMLU whose weights are currently available to the public.\\
Base your calculation on the date the open-weights model was first released to the public (including limited, semi-public releases), not the date when its weights were released.\\
Your answer should be a whole number, the number of calendar months between the first and second release:
e.g. if the first release was December 25, 1998 and the second was in June 1, 1999, your answer would be 6.\\
The value should be in the following unit: months.\\
\\
It is unlikely that you'll be able to find the number explicitly stated in a source; instead, you'll need to calculate it from numbers that you are able to find.\\
Your response must be calculated using an exact formula, where each variable is fixed as a number that can be traced to a source.\\
Nowhere should you estimate or guess; all inputs to your calculation must be found in sources.\\
If you are unable to figure out the number, set the success flag to false in your response.}

\subsection{Gather Evidence}

\exampletaskbox{Impact of LLMs on the environment}{Please gather all key items of evidence to help answer the following query:
What is the impact of LLMs upon the environment?
You must return 20 items.
The evidence you include should contain all necessary context such that a reader can fully understand the relevance, meaning, and implications of each item.}

\subsection{Populate Reference Class}

\exampletaskbox{List of AI datacenters}{Compile a list of all major AI datacenters in the US that are either operational, under construction, or planned.
An AI datacenter qualifies as major if it either hosts (or will host) 100,000 or more AI chips, or draws (or will draw) 1GW or more power at peak capacity.

For each datacenter you find, give a one-sentence description, enough to uniquely identify it.}

\subsection{Compile Dataset}

\exampletaskbox{US Software Developer Jobs 2019-2023}{Please compile a dataset fitting the following description:

Number of software developer jobs in the United States from 2019 through 2023 with the following column names: year, number of software developers, source, url, percent change. The percent change should be rounded to one decimal place.
\\
\\
    The dataset must have the following columns:
\\
    Column Name: year\\
    Description: The year for which the data is reported\\
    Type: number\\
\\
    Column Name: number\_of\_software\_developers\\
    Description: The number of software developers in the year\\
    Type: number\\
\\
    Column Name: source\\
    Description: The name of the source of the data\\
    Type: text\\
\\
    Column Name: url\\
    Description: The URL of the source of the data\\
    Type: text\\
\\
    Column Name: percent\_change\\
    Description: The percent change in the number of software developers from the previous year\\
    Type: number
}

\end{document}